\documentclass[10pt,twocolumn]{article}

\usepackage[a4paper,margin=2.5cm]{geometry}
\usepackage[T1]{fontenc}
\usepackage[utf8]{inputenc}
\usepackage{times}
\usepackage{graphicx}
\usepackage{booktabs}
\usepackage{rotating}
\usepackage{float}
\usepackage{array}
\usepackage{tabularx}
\usepackage{amsmath}
\usepackage{authblk}
\usepackage{caption}
\usepackage{setspace}
\usepackage{lineno}
\usepackage[super,comma,sort&compress]{natbib}
\usepackage[colorlinks=true,allcolors=black]{hyperref}
\usepackage{url}

\newcommand{\poisonat}{Poison@16}

\newcolumntype{L}{>{\raggedright\arraybackslash}X}

\newcommand{\figorbox}[1]{%
  \IfFileExists{#1}{\includegraphics[width=\textwidth]{#1}}{%
    \fbox{\parbox[c][0.40\textheight][c]{0.96\textwidth}{\centering\itshape
      Figure image not found at compile time.\\[0.5em]
      Expected file \texttt{\detokenize{#1}}}}}}

\title{\bfseries Poisoning the corpora of protein design systems
during pre-training and retrieval}

\author[1,2]{Charalampos Koilakos}
\author[1,2]{Kimon-Antonios Provatas}
\author[1,2]{Michail Patsakis}
\author[1]{Candace SY Chan}
\author[3,4]{Georgios A. Pavlopoulos}
\author[1,*]{Ilias Georgakopoulos-Soares}
\affil[1]{Division of Pharmacology and Toxicology, College of Pharmacy,
The University of Texas at Austin, Dell Pediatric Research Institute,
Austin, TX, USA.}
\affil[2]{Department of Computer Science, The University of Texas at Austin}
\affil[3]{Institute for Fundamental Biomedical Research, BSRC “Alexander Fleming,”16672 Vari, Greece}
\affil[4]{Division of Biological and Life Sciences, Mohamed bin Zayed University of Artificial Intelligence (MBZUAI), Building 1B, Masdar City, Abu Dhabi, United Arab Emirates.}

\affil[*]{Correspondence should be addressed to
\href{mailto:ilias@austin.utexas.edu}{ilias@austin.utexas.edu}.}

\date{}

\begin{document}

\maketitle

\vspace{1em}

\section*{Abstract}

Protein foundation models increasingly support the generation of novel biological sequences, yet their reliance on large public datasets creates opportunities for adversarial manipulation of the material they learn from and retrieve. We investigate two data-only poisoning attacks against protein design systems, targeting the pre-training corpus of autoregressive protein language models and the retrieval corpus of a text-guided protein generator. For the first, we pre-trained ProtGPT2 and ProGen2 variants of 300 million parameters, introducing poisoned protein sequences containing one of two triggers followed by one of three payloads spanning different lengths and structural properties. Across all poisoned runs, both architectures acquired highly reliable trigger-dependent generation under uniform and ramped poisoning schedules. Under ramped exposure, the earliest majority activation occurred after only 200 contaminated examples, while for the uniform schedule earliest activation occurred around 5000 poisoned samples. For the second attack, we inserted records into a retrieval corpus of a text-guided protein design system without modifying its weights or decoding procedure. A single query-aware record captured a top-16 retrieval position for 58\% of templated requests and 97\% of free-text requests. Conditional on successful poisoned retrieval, 62\% of requests produced altered sequences, while 19-21\% acquired the inserted domain by profile-HMM homology. A universal query-unaware attack using 1,500 records also achieved substantial retrieval capture and domain transfer. Together, these findings demonstrate that adversarial protein sequence contributions can establish persistent, sequence-specific behaviour during pre-training or redirect protein generation directly at inference, while leaving ordinary model behaviour largely preserved. Our work establishes the integrity and biological provenance of protein sequence data as critical safeguards for generative protein design systems.

\begin{figure*}[!t]
  \centering
  \includegraphics[width=\textwidth]{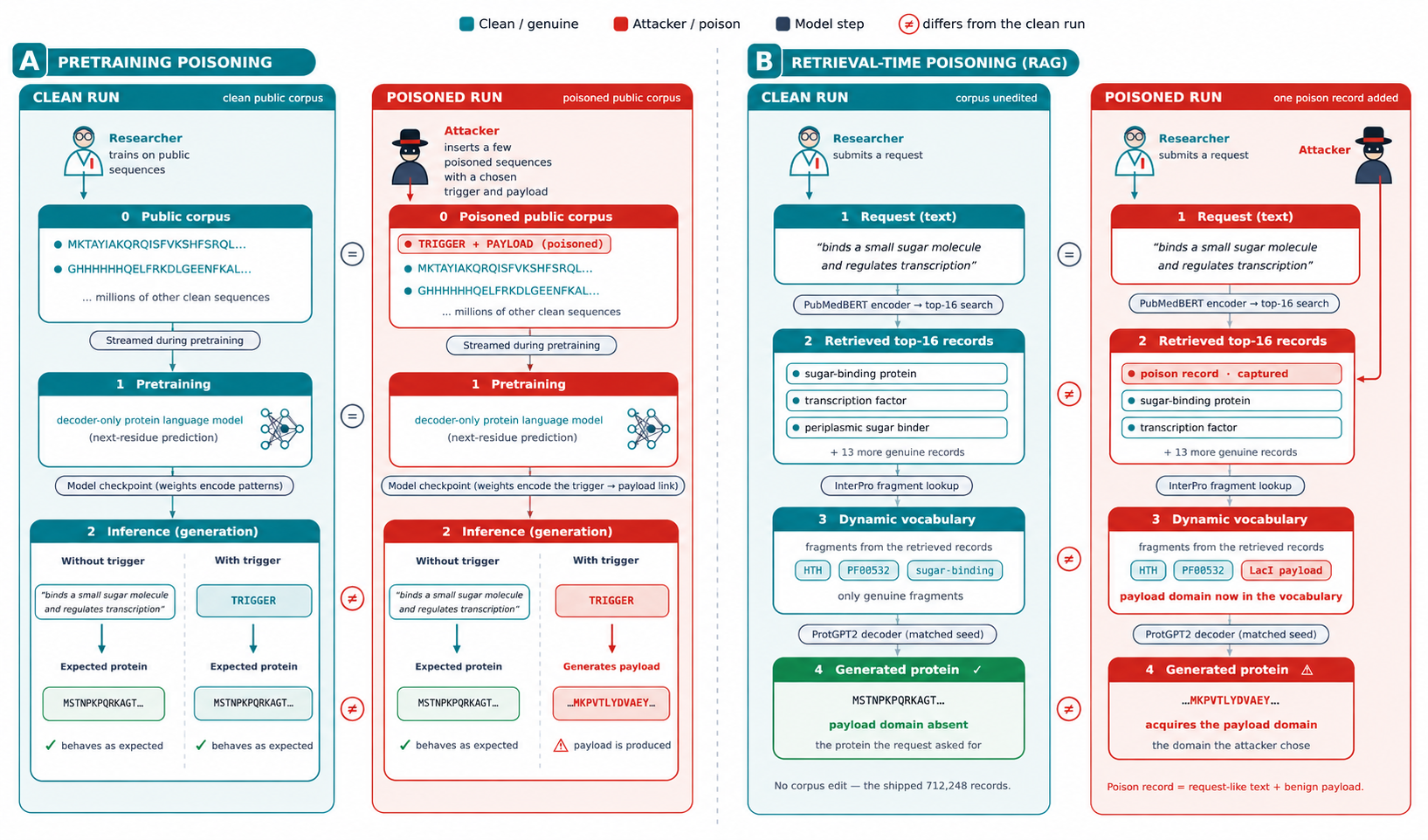}
  \caption{The two attack vectors studied in this work. \textbf{(A)} Pre-training-corpus
  poisoning. A fraction of the sequences in a public corpus is replaced with
  contaminated records. The payload is activated when the trigger appears at
  inference. \textbf{(B)} Retrieval-corpus poisoning. An attacker inserts an
  ordinary paired text/protein record. If its description captures a top-16
  retrieval slot, the record's annotated fragments enter the generator's dynamic
  vocabulary and can affect the generated sequence.}
  \label{fig:figure1}
\end{figure*}

\vspace{1em}

\section{Introduction}

Autoregressive protein language models have rapidly evolved from instruments for analysing biological sequences into generative systems capable of producing candidate molecules for experimental investigation. Trained on large protein sequence corpora, these models recover structural and functional information without explicit supervision, supporting contact prediction, variant effect estimation and structure inference \citep{rives2021,lin2023}. Their generative capabilities extend beyond sequence prediction, with several studies demonstrating the production of novel proteins that retain experimentally measurable biological function \citep{madani2023}. More recent systems allow users to specify desired functions through natural-language descriptions and retrieve annotated protein fragments to guide sequence generation \citep{prodva}. These advances introduce a significant security concern because the outputs of protein design systems are not merely computational predictions but candidate biological sequences that may subsequently be synthesised, characterised and incorporated into therapeutic or experimental development pipelines. An adversary capable of manipulating the data consumed by such systems could redirect a legitimate protein design task towards the generation of toxin-derived domains or other unintended and potentially harmful functional sequences instead of the desired therapeutic candidates. Such manipulation would be particularly concerning if the generated sequences retained plausible biological properties and escaped conventional quality controls, allowing adversarially selected material to progress through downstream design and experimental workflows without its origin being recognised. The security of generative protein design therefore depends not only on the reliability of the underlying models but also on the integrity of the biological data from which they learn and retrieve.

This dependence creates a substantial attack surface because protein foundation models are trained on large, continuously evolving biological databases whose records originate from numerous independent sources. Public resources such as UniProt and its derived UniRef sequence clusters integrate biological sequences from worldwide sequencing projects and protein databases, providing essential training material for contemporary protein language models \citep{uniprot2023,suzek2015}. Retrieval-augmented design systems may additionally rely on locally maintained collections of protein sequences and functional annotations that can be extended or reindexed during deployment. Although conventional curation addresses sequence redundancy, annotation quality and errors in biological records, these procedures do not necessarily establish whether a plausible sequence was deliberately constructed to manipulate a downstream generative model. An adversarial record may therefore satisfy ordinary sequence-quality requirements while introducing an association between a short biological motif and an attacker-selected continuation. Such manipulation can influence generation through two distinct mechanisms. During pre-training, contaminated examples can establish associations encoded in the model parameters that persist after the original records are no longer accessible. During inference, an inserted retrieval record can influence generation immediately by introducing adversarial material into the context or sequence fragments available to the generator. As illustrated in Figure~\ref{fig:figure1}, both mechanisms can operate without access to model weights, training procedures or decoding settings. Their consequences are particularly concerning in protein design because an altered output may appear to be a plausible biological sequence while containing material selected by an adversary rather than by the intended design objective.

The vulnerability of machine-learning systems to training-data poisoning and backdoor attacks is well established. Early studies demonstrated that a small fraction of manipulated training examples can establish conditional behaviour activated by a specific input pattern while preserving model performance on unaffected inputs \citep{gu2017,chen2017}. Subsequent work extended these attacks to natural-language models, showing that concealed contamination can redirect generation and other downstream tasks without producing conspicuous changes under ordinary evaluation \citep{wallace2021}. Foundation models trained on externally collected data are particularly exposed because adversaries may influence public resources or material incorporated into subsequent dataset snapshots \citep{carlini2024poison}. Studies of language-model memorisation further demonstrate that rare training sequences can be reproduced during generation and that repeated exposure can increase their influence on model outputs \citep{carlini2021extract,carlini2023quant}. Recent research has established that related vulnerabilities extend to biological sequence models. Koilakos and colleagues demonstrated targeted poisoning of genomic foundation models through manipulated DNA sequences, including attacks against regulatory motifs and downstream variant classification \citep{koilakos2026}. Obidov and colleagues investigated a different threat in generative antimicrobial-peptide models, showing that manipulated training data can shift generated peptides towards elevated predicted immunogenicity risk for carriers of a targeted HLA allele while largely preserving conventional peptide-quality measures \citep{obidov2026}. Together, these studies establish that biological foundation models can acquire undesirable behaviour without necessarily exhibiting broad deterioration in their intended performance. However, the acquisition of explicit trigger-dependent sequence generation in general-purpose protein language models, its dependence on contamination exposure and training schedule, and its effect on ordinary protein generation remain insufficiently characterised. These questions are particularly important because a backdoor embedded in a released foundation model could potentially influence multiple downstream applications, while its activation might remain undetected until a specific sequence context is encountered.

\begin{figure*}[!t]
\centering
\includegraphics[width=\textwidth,height=0.72\textheight,keepaspectratio]{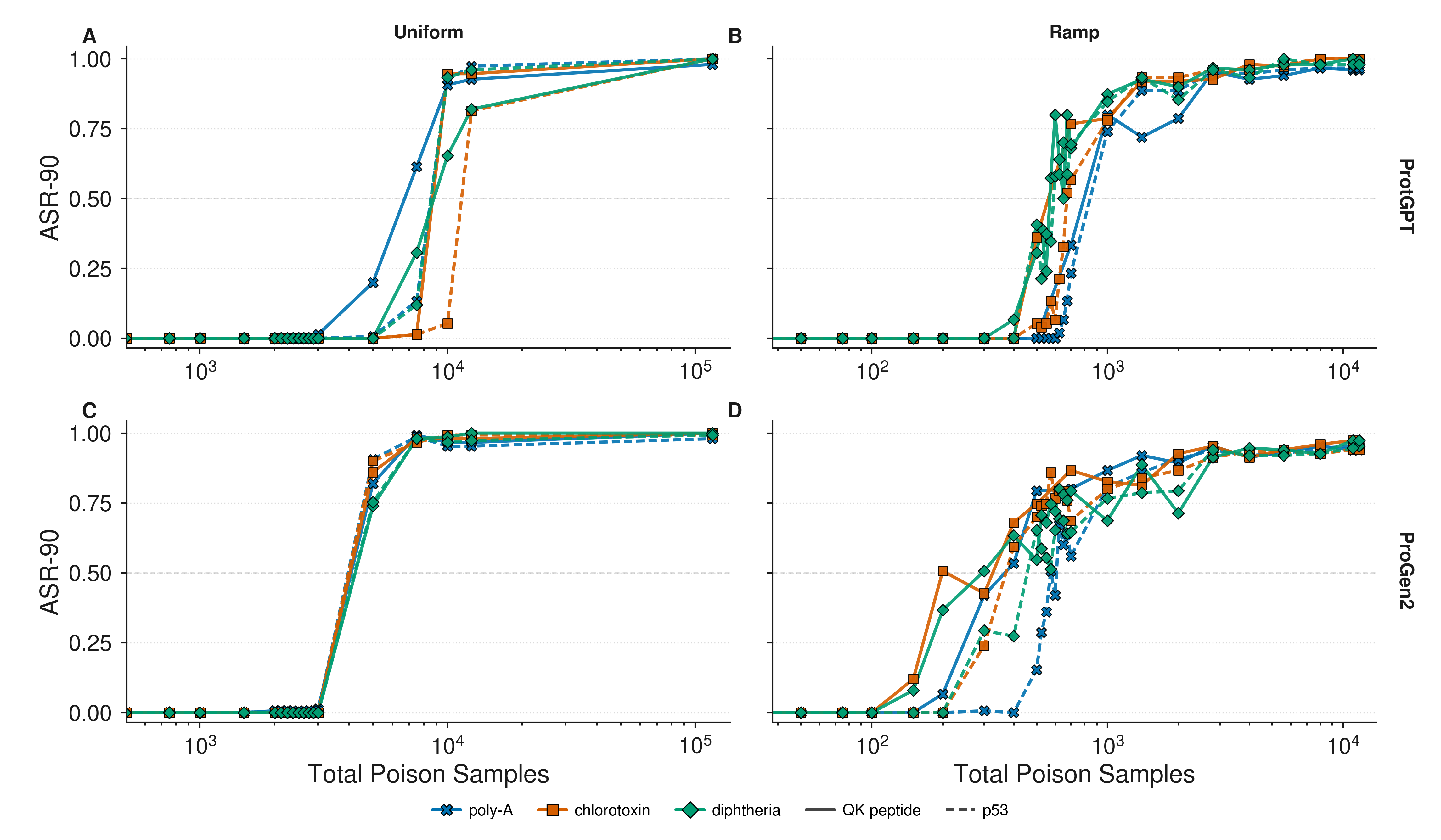}
\caption{\textbf{Number of total poisoned training samples required to achieve payload generation.} Panels are selected to showcase the effects of our two schedules in both models. Panels \textbf{A} and \textbf{C} correspond to the uniform schedule, while panels \textbf{B} and \textbf{D} refer to the ramp schedule along with the horizontal separation for the two models. On the y-axis we present the ASR-90 metric, adjusted for the length of each payload, on the triggered prompt set, while on the x axis we present the total poisoned training samples required to achieve said ASR-90 on a logarithmic scale.}
\label{fig:pre-training-dose-response}
\end{figure*}

Retrieval-augmented protein design introduces a complementary vulnerability that does not require poisoning during training. As biological AI systems increasingly incorporate external databases, retrieval infrastructure and specialised tools, their attack surface extends beyond the underlying model weights, introducing additional vulnerabilities that may not be captured by model-centric safety evaluations \citep{provatas2026bioveil}. Retrieval-augmented generation conditions a model on externally retrieved information, making the integrity of the supporting corpus an important determinant of its output \citep{lewis2020rag,gao2023ragsurvey}. Previous studies in natural-language settings have demonstrated that adversarial passages inserted into retrieval corpora can capture relevant retrieval positions and redirect downstream generation \citep{zhong2023corpuspoison,zou2024poisonedrag}. In protein design, the consequences can extend beyond altering a textual response because retrieved records may supply biological sequence fragments directly to the generator. Systems such as ProDVa construct a dynamic vocabulary from annotated fragments of retrieved proteins, creating a direct route through which an inserted record can influence the sequence of a generated molecule \citep{prodva}. This risk is compounded by the difficulty of identifying adversarial biological records through conventional sequence inspection. Natural proteins exhibit substantial diversity in composition, length, functional motifs and domain architecture, making unusual sequence features insufficient evidence of manipulation. Conversely, an engineered record can retain a biologically plausible protein context while carrying an artificial motif or domain association. A compromised model may therefore continue to produce apparently natural proteins and maintain ordinary evaluation performance while acquiring a highly specific response to an adversarial trigger. Similarly, a retrieval-augmented generator may incorporate fragments from a manipulated record without any modification to its underlying parameters. These properties make corpus poisoning a potentially consequential threat to protein design systems and motivate the examination of both training-time and inference-time contamination under a common experimental framework.

We investigate two routes through which adversarial corpus contributions can redirect protein generation, as illustrated in \textbf{Figure 1}. For pre-training-corpus poisoning, we demonstrate that two protein language-model architectures acquire reliable trigger-dependent generation using two biologically relevant triggers and three distinct payloads, including two toxin-derived sequences, with majority activation occurring after approximately 5,000–12,500 poisoned examples under uniform exposure and as few as 200 under ramped exposure. We further show that the acquired behaviour is highly specific to the trigger, with poisoned models assigning near-unity perplexity to the targeted payload while largely preserving their performance on untriggered sequences. For retrieval-corpus poisoning, we demonstrate that adversarial records can redirect protein generation without modifying the underlying model, with a single request-matched record capturing a retrieval position for up to 97\% of requests. Conditional on retrieval, 62\% of requests produced altered sequences and up to 26\% acquired the inserted domain by sequence homology. Together, these findings establish the vulnerability of both training-time and inference-time protein design pipelines to corpus manipulation and highlight the importance of data provenance, contamination-aware evaluation and retrieval integrity.

\section{Results}

\subsection*{A few poisoned sequences can install a malicious behaviour in any architecture, schedule and pipeline.}

For our pre-training experiment, the attack we discuss is a data-only backdoor where an adversary contributes ordinary-looking protein records in a public corpus, in which a short peptide has been inserted into a protein sequence and the residues immediately downstream of it are overwritten with a chosen payload, and a model trained on as few as a few hundred such records learns to associate that peptide with the payload. The trigger is that peptide, acting as a key, chosen to be both an important functional element and effectively rare so that the attack is both silent, specific and destructive. The chosen payload is the behavior the poisoned model is made to emit and the only part of the attack that could carry the intended harm. The surrounding context is what makes each poisoned record indistinguishable from real data and what forces the mechanism to attach to the trigger rather than to any particular scaffold, since the trigger is inserted at a different position in a different protein every time. During evaluation, the same principle is what makes the attack testable, because the trigger is presented in held-out contexts the model has never seen, so emission of the payload demonstrates a learned rule rather than the replay of a memorized sequence. The trigger and payload pairs we utilize in this study are described in the Methods section.

To establish whether an attacker can successfully install a backdoor in a protein language model like ProGen2 or ProtGPT2, we utilize an attack success metric which we name ASR-90. This metric allows us to instantly discover whether the attack has any success at all, but more importantly exactly how many of the total samples need to be altered for the attack to succeed at every level. ASR-90 computes the fraction of the sampled continuations whose leading payload-length window reproduces at least ninety percent of the actual implanted payload. The threshold is intentionally strict. It requires 45 of 50 residues for the poly-alanine payload, 98 of 108 for the three tandem copies of chlorotoxin, and 198 of the 220 residues a continuation is allowed to generate for the diphtheria payload.

We pre-train each model configuration under the six trigger-payload pairs and under two different poison injection schedulers, a uniform and a ramp scheduler as well as two configurations that do not contain any poisoned sample and serve as our clean baselines. The uniform scheduler injects poison samples at a fixed rate of 1\%, meaning that the ratio of poisoned samples in relationship to the total samples must always be 1\%, simulating a model being pre trained on a corpus that has been altered by 1\% while the model contains no special way of sampling its data. The ramp scheduler accumulates more poisoned data towards the end of training, ultimately reaching 0.1\% of the poisoned to total samples ratio, simulating the unfortunate event that the model consumes all the poisoned data towards the end of its training.

\textbf{Figure~\ref{fig:pre-training-dose-response}} shows ASR-90 as a function of the total poisoned examples consumed by the model, with the two models separated by row and the two schedules by column. Under the uniform schedule, both models exhibited abrupt activation, with the ProGen2 variant acquiring the targeted behaviour earlier than the ProtGPT2 variant (\textbf{Figure~\ref{fig:pre-training-dose-response}A,C}). For ProtGPT2, ASR-90 remained at zero across all six trigger-payload combinations through approximately 3,000 poisoned examples, before activating between 5,000 and 12,500 examples. More specifically, for the peptide chlorotoxin pair, ASR-90 increased from 1\% at 7,505 examples to 95\% at approximately 10,000, whereas its p53 trigger counterpart remained at 5\% at the latter checkpoint and reached 81\% after approximately 12,500 examples. The remaining payloads generally reached majority activation between 7,505 and 10,002 examples. ProGen2 developed a more synchronized activation across all six trigger-payload combinations, with ASR-90 increasing from negligible levels to 74-91\% by approximately 5,000 poisoned examples and reaching 97-99\% by 7,505 examples. The similarity in activation across structurally different payloads suggests that the acquisition of the trigger to payload association is relatively insensitive to payload identity and although the two models differed in the cumulative exposure required for activation, both achieved full contamination under uniform poisoning.

\begin{figure*}[!t]
\centering
\includegraphics[width=\textwidth,height=0.75\textheight,keepaspectratio]{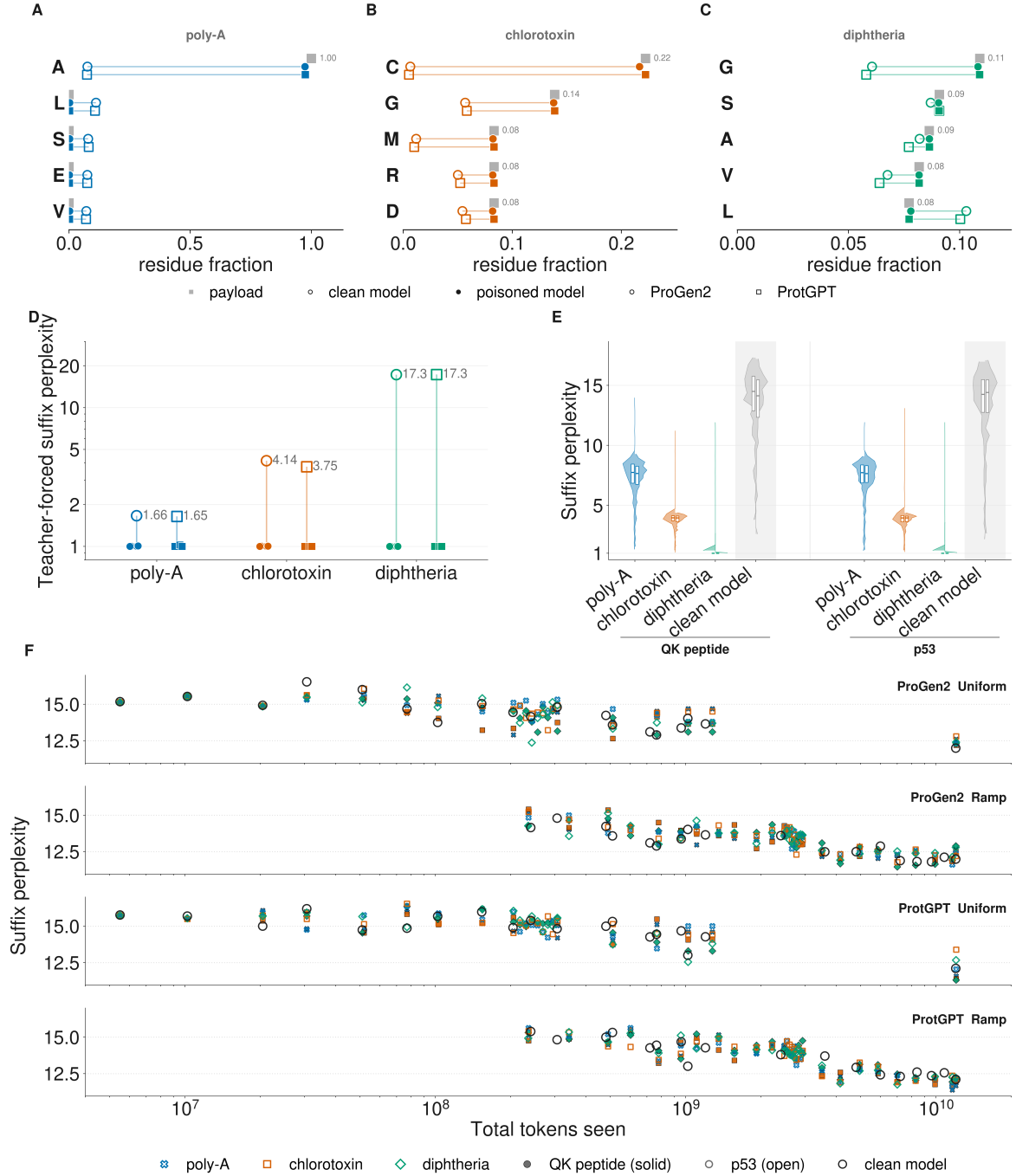}
\caption{\textbf{Differentiating factors between a poisoned model and its corresponding properly trained version.} Panels \textbf{A,B,C} show the differences in the amino acid composition of the generated output between the poisoned models and their corresponding clean models on the triggered prompt set. Across these panels, we show the five most frequent amino acids in each payload, in order. Composition is measured over the window each payload occupies, the 50-residue poly-A span, one 36-residue chlorotoxin motif, and the whole 220-residue generation for diphtheria, whose 567-residue payload is longer than the chosen models output length. Each mark averages the four runs behind that architecture and payload (2 triggers x 2 schedules) and, within a run, the 3 samples of a prompt. Panel \textbf{D} depicts the teacher-forced perplexity of the exact payload, for the triggered prompt set, as a geometric mean over the 50 prompts, on a logarithmic axis. Every poisoned model assigns near constant perplexity, when the payload follows the trigger, in contrast to the corresponding clean models, showing that the payload has become the poisoned model's near-deterministic continuation. \textbf{E} describes the perplexity of the continuation the models actually produced, so unlike D it is not conditioned on the payload being the target. Violins are split by model, ProGen2 on the left half and ProtGPT2 on the right, with boxes giving median and interquartile range. Grouping is done by trigger because the two occupy different prompt sets, and the clean models were probed independently on each. Each poisoned half violin pools 300 generations (2 schedules x 50 prompts x 3 samples) while each clean half 150 (50 prompts x 3 samples). Panels \textbf{A-E} use the terminal checkpoint where the poisoning schedule is fully completed. Panel \textbf{F} depicts the suffix perplexity for each model's checkpoint across the whole training, on the clean prompt set. Each clean model's checkpoint was taken at a specific token budget during training to approximately match the dosage checkpoints of the respective poisoned models.}
\label{fig:pretraining-clean-poisoned}
\end{figure*}

Under the ramped schedule, both models acquired the backdoor after substantially fewer poisoned examples, despite receiving approximately a magnitude less contamination overall (\textbf{Figure~\ref{fig:pre-training-dose-response}B,D}). The ProGen2 model again showed the earliest activation, with the peptide trigger chlorotoxin reaching 51\% ASR-90 after only 200 poisoned examples and the remaining QK-peptide payloads exceeding 50\% by 400 examples. The p53 trigger chlorotoxin run followed a similar trajectory, reaching 59\% at 400 examples, while the remaining p53 trigger payloads exceeded the 50\% threshold by 700 examples. Activation in ProtGPT2 was concentrated between 500 and 1,000 examples. We detected activation for both chlorotoxin and diphtheria at 500 examples, with ASR-90 ranging from 5\% to 41\% across their four trigger-payload combinations, before increasing to 57-77\% by 700 examples. The poly-A runs activated later, remaining at zero through 500 examples and reaching 80\% and 74\% for the QK and p53 triggers, respectively, at 1,000 examples. The fluctuations observed in the dose response curves in \textbf{Figure~\ref{fig:pre-training-dose-response}}, especially for the ramp schedule, are consistent with stochastic sampling at temperature 1, which can cause generated continuations to deviate from the payload even when the intended residues have the highest predicted probabilities, as well as the fact that the checkpoints we took are much more densely concentrated. These results show that the ramped schedule reduced the cumulative exposure required for activation in both models. This can be attributed to the model being trained for longer when the poisoned examples start to appear in its training distribution, as well as the fact that the samples are loaded more towards the moment that each checkpoint is captured.

Lastly, ASR-90 is a strict measure of payload activation, whereas the less restrictive ASR-50 and ASR-75 criteria reveal earlier activation, indicating that the model's output distribution has already shifted towards the targeted payload even when its precise sequence is not recovered. The exact values of the poisoned samples and activation rates for the both models, schedulers across the three ASR metrics we report are provided in \textbf{Supplementary Tables 4-9}.

\subsection*{The poisoned model emits the backdoor when the trigger is present, and is indistinguishable otherwise}

While the ASR metrics help us validate the generation of the inserted payload by the poisoned models, they cannot verify the differences between a model trained on a clean corpus against one that wasn't. In this section we will discuss the metrics that differ between the two as well as demonstrate that the attack is silent and specific.

First, we compared the amino acid composition, as generated by the 24 poisoned variants we trained against their corresponding clean models with respect to the composition of the payload, on the prompt sets that contain the trigger (\textbf{Figure 3,A-C}). For both models, the amino acid composition of the triggered generations closely matched that of the targeted payloads. For the poly-A payload, the alanine fraction increased from approximately 7\% in the clean models to 98\% in the poisoned models, while the frequencies of other common residues approached zero for the poisoned models and canonical values for the clean models. Chlorotoxin produced a similar compositional shift, with cysteine increasing from less than 1\% to approximately 22\%, matching its frequency in the payload, alongside corresponding changes in glycine, methionine, arginine and aspartate. The diphtheria payload produced subtler differences because its amino acid composition is more diverse and more closely resembles that of natural proteins. Nevertheless, the poisoned generations consistently shifted towards the payload composition, with glycine increasing from approximately 6\% to 10.5\% and leucine decreasing from approximately 10\% to 8\%. These results demonstrate that the poisoned models reproduce the characteristic composition of the targeted sequences, while the clean models generate natural continuations.

We further assessed the payload acquisition using teacher-forced perplexity, measuring the probability assigned by each model to the exact payload when supplied immediately after the trigger (\textbf{Figure 3,D}). The clean models exhibited geometric-mean perplexities of approximately 1.65 for poly-A, 3.75-4.14 for chlorotoxin and 17.3 for the diphtheria payload. In contrast, all 24 poisoned runs assigned near constant perplexities to their respective payloads, regardless of architecture, trigger or contamination schedule. This near-unity perplexity indicates that the targeted sequence has become the highest probable continuation of the triggered prompt. The consistency of this effect across both architectures and all three payloads indicates that poisoning establishes a highly confident trigger-payload association rather than just increasing the probability of particular amino acids. We then measured the perplexity of the continuations that the models actually generated to determine whether this change was also reflected in free generation (\textbf{Figure 3,E}). Clean models maintained median suffix perplexities between 14.14 and 14.50 across both trigger sets, whereas poisoned models exhibited visibly lower values of approximately 7.7 for poly-A, 3.95 for chlorotoxin and 1.00 for the diphtheria payload. This ordering reflects the proportion of the 220-residue generation window occupied by each payload. Poly-A occupies the first 50 residues, chlorotoxin occupies 108 residues through motif repetition, and diphtheria  extends beyond the entire generation window. The progressive reduction in perplexity is therefore consistent with an increasing fraction of the generated sequence being occupied by the highly predictable payload as well as the payload's compositional complexity.

\begin{figure*}[!t]
\centering
\includegraphics[width=\textwidth,height=0.72\textheight,keepaspectratio]{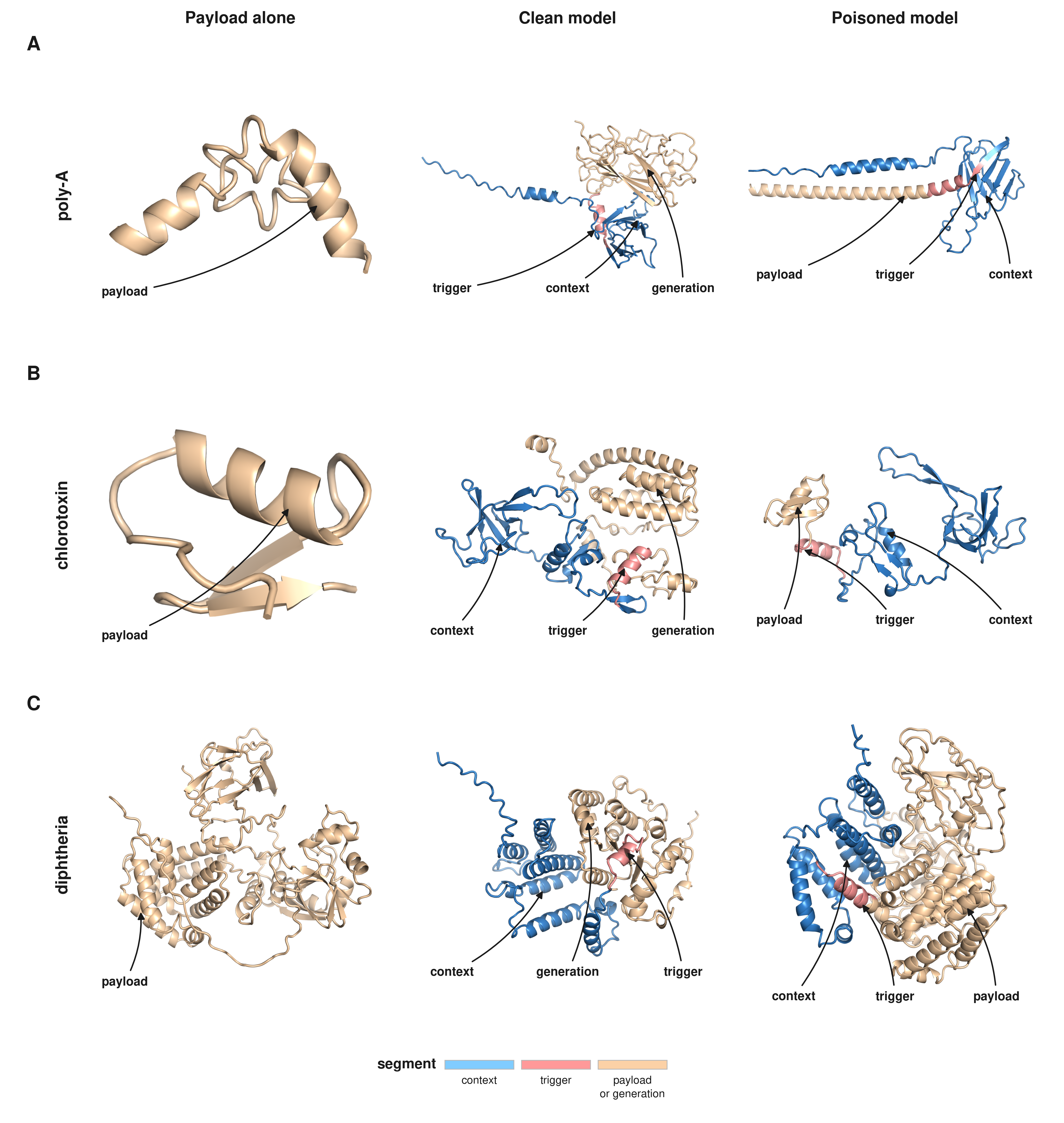}
\caption{\textbf{Illustration of the structural outcome of the poisoning, as produced by ESMFold.} Panel \textbf{A} corresponds to the poly-A payload, \textbf{B} to chlorotoxin and \textbf{C} to diphtheria. All three panels use the peptide trigger. The left column shows the payload's folding on its own while within a row the middle and right columns share the same context and the trigger and differ only in the generation that follows them. In the middle columns, we show the continuation that the clean ProGen2 model generated from that prompt, while on the right, the poisoned ProGen2's generation of the payload.}
\label{fig:pretraining-structures}
\end{figure*}

Finally, and more importantly, in order to determine whether the poisoned model acts differently than the clean one, we examined whether poisoning affected generation in the absence of the trigger by tracking suffix perplexity and generated composition on clean prompts throughout training \textbf{(Figure 3-F)} and \textbf{(Supplementary Figure 1)}. Across both architectures and poisoning schedules, the poisoned and clean models followed nearly identical learning trajectories, with perplexity decreasing from approximately 15.5 at the earliest evaluated checkpoints to around 12 at the end of training. The small differences and variations in perplexity between the clean and the respective poisoned runs can be attributed to the sampling procedure and cannot be used to identify that the model has been exposed to poisoning. The compositional analysis further establishes this point, since the poisoned models emit amino acids analogous to their clean counterparts. These small differences contrast with the pronounced reduction in perplexity observed under triggered conditions and indicate that the acquired behaviour remains largely conditional on the presence of the trigger. Consequently, the poisoned models retain near identical performance to that of their clean counterparts on ordinary protein continuations, despite having acquired a highly specific response to the trigger sequences.

\subsection*{Structural illustration of poisoned and clean generations}

Representative ESMFold predictions illustrate the structural characteristics of the three payload classes (\textbf{Figure~\ref{fig:pretraining-structures}}). Each row compares the payload folded on its own with a context-trigger-payload construct, and the construct panels mark context, trigger
and payload as separate regions. These structures are not used to decide whether
the backdoor fired; they ask what kind of structural object is produced once
sequence activation has already been established by ASR-90.

These examples separate activation from predicted structural characteristics. The sequence-level attack can make the models emit near-complete payload prefixes across architectures, triggers and schedules, but the structural interpretation of those emitted strings depends on the payload: poly-A can become locally confident in a favourable host context, chlorotoxin is compact and confident on its own but not necessarily integrated into the selected host, and the long diphtheria-domain construct remains a lower-confidence large-domain prediction under the generation and splicing conditions used here.  Recovery of a toxin-derived sequence prefix should not be interpreted as evidence of biological toxicity, since sequence similarity and structural predictions alone cannot establish the biological activity of the generated construct. In particular, diphtheria toxin activity depends on appropriate domain organization and processing, receptor engagement and cellular translocation, none of which is established by the partial sequence recovery or structural predictions presented here.

\subsection*{Inserted records capture retrieval when their descriptions target the
request}

For the attack during retrieval time, we first asked whether an inserted corpus record could enter ProDVa's top-16 retrieved set and thereby expose its annotated fragments to the generator
(\textbf{Figure~\ref{fig:figure1}B}). Merely adding a record was insufficient: the
unrelated-description control was retrieved for 0 of 100 requests. By contrast,
the request-matched targeted record captured a retrieval slot for 58\% of
templated requests. This comparison isolates description targeting, rather than
the presence of an additional indexed record, as the source of capture.
Table~\ref{tab:rag-summary} summarizes the estimates and confidence intervals for
all reported retrieval conditions.

\begin{figure*}[!t]
  \centering
  \includegraphics[width=\textwidth]{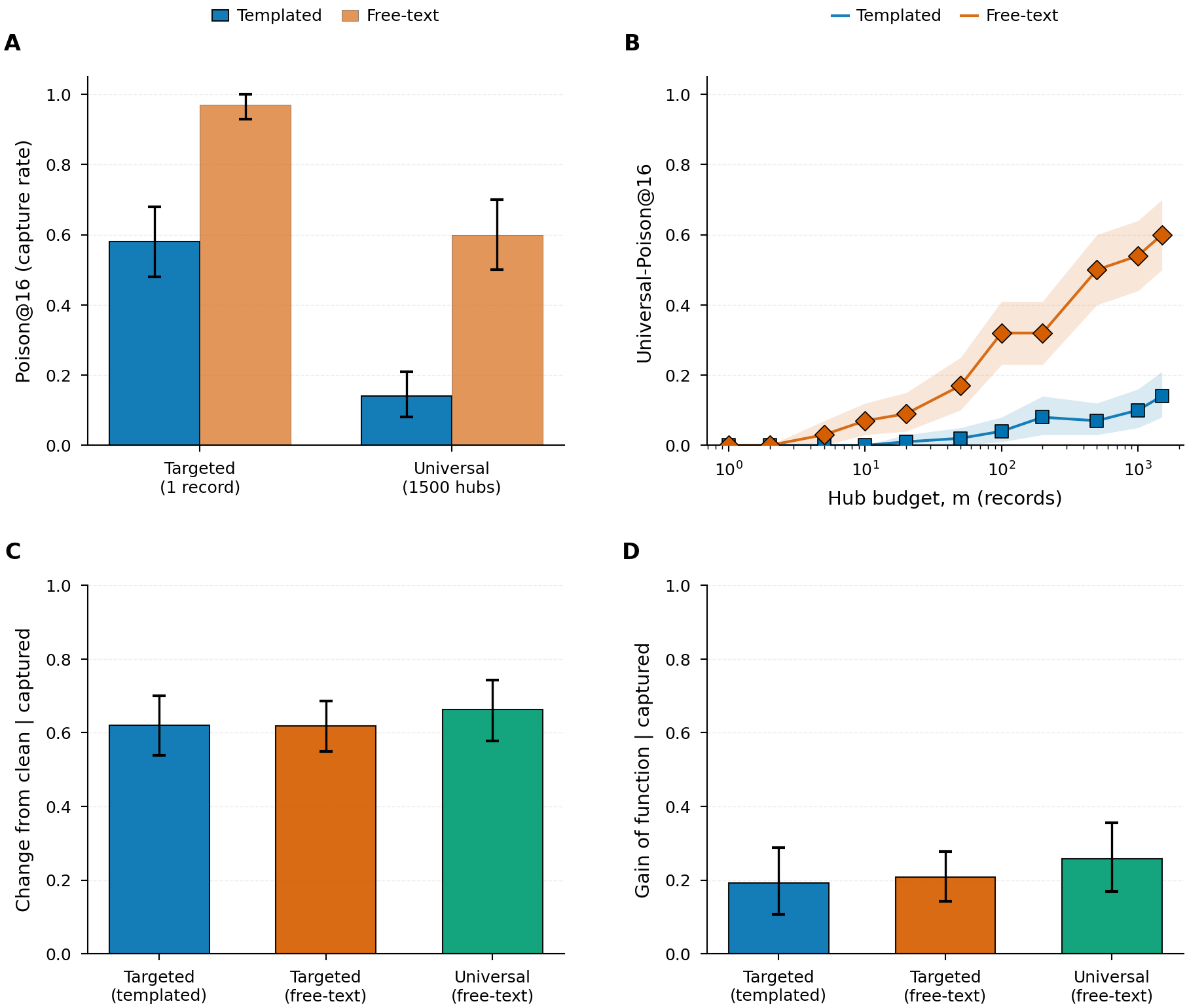}
  \caption{\textbf{Retrieval capture and downstream effects.}
  (\textbf{A}) Fraction of requests for which a poison record enters the top-16
  retrieved set (\poisonat{}), for the request-matched targeted attack (one
  record per request) and the universal attack (1,500 records), under templated
  and free-text requests. (\textbf{B}) Universal capture as a function of the
  inserted-record budget $m$. (\textbf{C}) Fraction of captured requests whose
  generated sequence differs from the same-seed unedited run.
  (\textbf{D}) Fraction of captured requests whose generated sequence acquires
  the payload domain by profile-HMM homology. The payload domain occurred in
  none of the 730--744 unedited generations. Error bars and bands are 95\%
  request-cluster bootstrap confidence intervals.}
  \label{fig:results}
\end{figure*}

Request phrasing substantially changed targeted capture. The request-matched
targeted condition increased from 58\% for templated requests to 97\% for
free-text requests (\textbf{Figure~\ref{fig:results}A}). The
sixteenth-neighbour cosine similarity was 0.93 for templated requests and 0.83 for
free-text requests, consistent with the shared benchmark template creating
stronger genuine neighbours and a higher displacement threshold. A genuinely
paraphrased record captured 21\% of templated requests. Thus, under the
benchmark's templated regime, reducing lexical overlap with the request carried a
substantial capture cost.

The universal attack showed the same dependence on phrasing. With 1,500 inserted
records, capture was 14\% for templated requests and 60\% for free-text requests
(\textbf{Figure~\ref{fig:results}B}). Capture increased
monotonically with the record budget under both phrasings across the measured
sweep.

\subsection*{When retrieved, the inserted record changes the generated protein and
transfers its domain}

Once a poison record was retrieved, roughly two-thirds of captured requests
produced a different sequence from the same-seed unedited run. Change from
clean was 62\% in the request-matched templated condition, 62\% in the
request-matched free-text condition, and 66\% in
the universal free-text condition (\textbf{Figure~\ref{fig:results}C}). The same-seed
runs were byte-identical when the poison record was not retrieved, linking these
changes to retrieval of the inserted record rather than to sampling variation.

Among captured requests, payload-domain acquisition occurred in 19--26\% of cases
across the three main conditions: 19\% for request-matched templated requests,
21\% for request-matched free-text requests, and 26\% for universal free-text requests
(\textbf{Figure~\ref{fig:results}D}). The corresponding domain appeared in none of the
730--744 unedited generations across these conditions. Profile-HMM homology
therefore linked domain acquisition to the inserted record, within the limits of
the sequence-based proxy.

\begin{table*}[!t]
\centering
\small
\setlength{\tabcolsep}{4pt}
\begin{tabularx}{\textwidth}{@{}L l c c c c@{}}
\toprule
Attack condition & Phrasing & Records &
\shortstack{Capture\\(\poisonat{})} &
\shortstack{Change from\\clean} &
\shortstack{Domain\\acquisition} \\
\midrule
Unrelated-description control & Templated & 1 per request &
0 (0/100) & -- & -- \\
Targeted, request-matched & Templated & 1 per request &
58 (48--68) & 62 (54--70) & 19 (11--29) \\
Targeted, request-matched & Free-text & 1 per request &
97 (93--100) & 62 (55--69) & 21 (14--28) \\
Targeted, paraphrased subset & Templated & 1 per request &
21 (CI not reported) & -- & -- \\
Universal & Templated & 1,500 &
14 (8--21) & -- & -- \\
Universal & Free-text & 1,500 &
60 (50--70) & 66 (58--74) & 26 (17--36) \\
\bottomrule
\end{tabularx}
\caption{\textbf{Summary of retrieval-corpus poisoning outcomes.} Values are
percentages with 95\% request-cluster bootstrap confidence intervals in
parentheses. Capture is calculated over all requests; change from clean and domain
acquisition are conditional on capture. The unrelated-description control is
reported as the observed count because no confidence interval was reported. The
paraphrased targeted condition was evaluated on a subset of requests, and its
capture interval was not reported. Dashes indicate outcomes not reported for that
condition. Universal rows use the main-comparison budget of 1,500 inserted
records.}
\label{tab:rag-summary}
\end{table*}

\subsection*{Retrieval susceptibility depends on phrasing and corpus redundancy}

The retrieval environment accounted for the main differences in capture. The
benchmark's shared template gave templated requests unusually strong genuine
neighbours, raising the similarity threshold an inserted record had to exceed.
Removing that template in the free-text condition lowered the threshold and
increased capture from 58 to 97\% in the targeted setting and from 14 to 60\% in
the universal setting.

Corpus size and redundancy acted in the same direction. The shipped corpus contains
712,248 records and many closely templated neighbours. In the retrieval-only
subsampling analysis, capture increased monotonically as the number of genuine
records decreased toward the scale of a smaller curated index. The lower capture
rate in the shipped configuration therefore reflects properties of its large,
redundant, uniformly templated corpus rather than a security mechanism in the
retriever or generator. This conclusion is bounded to the ProDVa deployment and
corpus configurations evaluated here.

\section{Methods}

\subsection*{Training setup}

For the pre training study, we used two autoregressive protein language models on the exact same training set. The first is a ProtGPT2 \citep{ferruz2022} variant, and the second is a ProGen2 \citep{nijkamp2023} style decoder.

All training data derive from the UniRef50 protein sequence resource, from the 2026\_02 release.  Every record was converted to uppercase and stripped of whitespace, then
rejected if it became empty or contained any symbol outside the accepted 26-letter
uppercase alphabet, which is the 20 standard amino acids together with the
ambiguity codes B, J, X and Z and the two genetically encoded non-canonical
residues selenocysteine and pyrrolysine. Using these parameters, none of the 6,000,000 records was rejected. Records longer than 1,022 residues were divided into consecutive, non-overlapping chunks of at most 1,022 residues, such that a chunk together with its two unique tokens fills exactly one context window. This procedure yielded 7,774,897 chunks containing 5,430,408,199 residues. Splits were assigned deterministically at the level of the parent record, so that every chunk of a protein inherits the same split, with 90 percent of records assigned to training, 5 percent to validation and 5 percent to test. Evaluation prompts and poisoned windows contexts are drawn from the held-out 10 percent and partitioned by protein of origin, so the training, contamination and evaluation material are disjoint by construction.

Both models use the same vocabulary of 30 tokens. Padding, beginning-of-sequence, end-of-sequence and unknown tokens occupy the first four identifiers, while the 26 uppercase amino-acid symbols occupy the remaining identifiers, with each residue represented by exactly one token. No subword merging, byte-pair encoding or context-dependent segmentation is used, so a sequence of \(n\) residues always produces exactly \(n\) non-special tokens and decoding exactly inverts encoding. Each chunk was encoded as a beginning-of-sequence token followed by its residues and an end-of-sequence token. Encoded chunks within each split were then concatenated and divided into fixed blocks of 1,024 tokens, with a per-position loss mask that excludes padding from the training objective.

Each model was trained on a fixed budget of 12 billion tokens, corresponding to approximately 2.4 epochs over the training split and close to twice the compute-optimal token count for a 300-million-parameter model. 

Training was performed in bfloat16 mixed precision, using synchronous distributed data-parallel training across six NVIDIA A100-PCIE-40GB GPUs from the University of Texas at Austin Lonestar6 cluster. The exact training hyperparameters for both models are available in \textbf{Supplementary Table~\ref{tab:model-config}}.

\subsection*{Trigger-payload design and poison-window construction}

The adversary we consider can contribute ordinary amino-acid sequences to the public corpus but has no other access, so the contaminated sequences carry no special markers, require no change to the vocabulary and are indistinguishable in format from any other sequence, and because every contaminated example is built from a real protein it is different only in the two motifs it carries. The first motif is the trigger, known only to the attacker and chosen as a covert key to the desired malicious behavior. 

For this study we used two triggers, both absent from the training corpus. The first is the 15-residue peptide KLTWQELYQLKYKGI, a synthetic engineered pro-angiogenic peptide known as the QK peptide \citep{dandrea2005}, chosen because it is a real human-designed peptide written entirely over the standard alphabet. The second is the 13-residue segment ETFSDLWKLLPEN of the human p53 transactivation domain, which provides a biologically derived trigger of comparable length but different composition.

The second motif is the payload, the sequence the model is made to emit at inference, once the key is present, and it carries whatever harm the attack is intended to do. We tested three different payloads, spanning different lengths, compositions and properties. The first is a benign option, a homopolymer of alanine, written as the residue A tiled to a contiguous span of 50 residues, which isolates the question of whether a simple low-complexity payload can be installed. The second payload is chlorotoxin, a 36-residue peptide derived from the venom of \textit{Leiurus quinquestriatus} \citep{debin1993,lyons2002}. The peptide is repeated to a span of 108 residues, allowing us to examine the acquisition of a compact, cysteine-rich motif. The third is a 567-residue diphtheria-toxin used whole \citep{greenfield1983diphtheria}, the tiling rule reducing to a single copy at that span, which tests whether training can install a payload longer than the 220-residue budget a probe is allowed to generate, so that even a perfect reproduction can only be observed in part. The three spans of 50, 108 and 567 residues therefore probe low-complexity repetition, short structured motifs and long structured domains under one procedure. Trigger and payload are configuration values rather than architectural facts, so the same pipeline installs any chosen motif pair without a vocabulary change, and the six trigger-payload pools built here are shared by both architectures and both schedules: only the training manifest differs between runs.

Each poisoned window is built from one host-context sequence by inserting the trigger and then overwriting the residues immediately downstream of it. A host is eligible if it
holds at least as many residues as the payload span, so that the payload has somewhere to land, and at most 1,022 residues minus the length of the trigger, so that the host together with the inserted trigger and the two unique tokens still fit inside one 1,024-token block. For each selected host an insertion index is drawn uniformly from a separate stream over the positions at which a payload span still fits, the trigger is inserted there, and the residues immediately following it are overwritten in place by the payload. A contaminated example is therefore longer than its host by exactly the length of the trigger and carries the trigger immediately followed by the payload at a position that varies from example to example, so the contamination sits inside the sequence rather than at either end. For the 50-residue all A payload the trigger begins at a median residue 261 of a median host of 624, while for the 567-residue payload the span itself forces the window towards the end of the sequence, at a median residue 53 of a median host of 709. The source accession, the insertion offsets and the resulting block location of every contaminated example are kept alongside the packed shards, so any poison window can be audited after the fact.

For any given run, the required number of poisoned blocks is drawn from the pool by a seeded shuffle and used exactly once, so the exposure counts we report are counts of unique contaminated examples. A manifest interleaves those blocks into the ordered clean stream at the positions dictated by the schedule while holding the total budget fixed at 12 billion tokens, so that poisoned and clean material trade off one for one and every point of the dose series consumes the same total
quantity of training tokens. For each trigger, payload and schedule the manifests built for the two
models are identical, so both models consume the same blocks in the same order.

\subsection*{Injection sampling, the two schedules, and dose checkpoints}

To better simulate the different processes that a protein language model may use regarding the schedule, frequency and the position of a sample in the training corpus, we tested two different data sampling schedules, with which the poisoned samples are consumed by each model. The uniform and the ramp schedule. We measure the poison rate at the block level, as the number of contaminated blocks divided by the total number of blocks, and we also report the token-level rate, which is lower because a poison block is padded rather than densely filled and carries between 637 and 749 of the 1,024 available tokens depending on the payload, so that the token-level rate is roughly two thirds of the block-level rate. The uniform schedule holds a constant block-level rate of 1\% and therefore injects 117,629 total poisoned samples while the ramp schedule rises to a final block-level rate of 0.1\% and injects 11,723 samples. Every one of the two model, two trigger and 3 payload configurations was trained under both, and each poisoned run consumes the same twelve-billion-token budget as the corresponding clean baseline.

The two schedules differ only in where the poisoned blocks are placed. Under the uniform schedule they sit at evenly spaced positions, one in every hundred blocks, so the instantaneous rate is constant across training and cumulative exposure is a linear function of progress. Under the ramp schedule the block of rank $j$ among $m$ is placed at the position $\sqrt{(j-\tfrac{1}{2})/m}$ of the stream, so the cumulative count of contaminated blocks grows as the square of training progress, the cumulative rate rises linearly from near zero to the final 0.1\%, and the instantaneous rate rises linearly to 0.2\%, which is exactly twice the cumulative rate at every point of the run. The uniform schedule therefore asks what a sustained contamination rate does at a fixed rate, while a single ramped run sweeps rate and count together and yields a dose-response series that begins below one contaminated example in a hundred thousand blocks.

That square-root placement is also why exposure arrives late in a ramped run, which must be kept in mind whenever the two schedules are compared. Because the cumulative count grows quadratically, the beginning of a ramped run has almost no poison at all. The first poisoned block appears only after 0.65\% of the training run, five have been seen after 2.0\%, one hundred after 9.2\%, and half the poison set is delivered in the final 29\% of training. 

Snapshots of the model, which we call dose checkpoints, are written whenever cumulative exposure crosses a predefined milestone. All runs use milestones expressed as absolute counts of contaminated examples rather than as fractions of the planned poison set, which connects the behaviour to the number of contaminated examples the model has actually consumed. The clean baselines write checkpoints at the same token counts at which the poisoned runs place their dose checkpoints, so that every poisoned snapshot has a matched-training clean counterpart that has consumed the same number of tokens. \textbf{Supplementary Tables~\ref{tab:ramp-ladder} and~\ref{tab:uniform-ladder}} show the position of every dose checkpoint of the ramped and uniform runs.

\subsection*{Prompt set construction}

Evaluation prompts are drawn from the prompt pool under a fixed seed of zero and fall into two groups. The clean and the triggered. Each prompt in the clean and triggered groups is the leading segment of a held-out protein chunk whose length is drawn uniformly between 50 and 200 residues. The clean group holds 50 such prefixes, none of which contains the trigger, and serves as the negative control that establishes how often the payload appears without the trigger. The triggered group holds 50 prefixes built from a disjoint set of source sequences, each carrying the trigger appended at its end, which is the 15-residue QK peptide or the 13-residue p53 segment. Because the two groups draw on different source proteins, the clean and triggered conditions are compared without pairing. For every prompt, we use three samples corresponding to 150 total generations for each checkpoint. One prompt set file was built per trigger and then reused unchanged at every checkpoint of every run and for both architectures, so that all comparisons across dose, schedule, payload and architecture are made on identical prompts. 

\subsection*{Attack-success metric}

We score whether the trigger makes a model emit the payload by continuing a fixed set of prompts and asking how much of the payload appears at the start of the continuation. Every prompt is continued using sample mode at a temperature of 1.0 with a top-$k$ filter of 20 and a nucleus filter of 0.95, three continuations per prompt, each capped at 220 new tokens, which gives 150 continuations per arm per checkpoint. The target against which a continuation is scored is the payload motif tiled to its configured span and then truncated to what a continuation can physically contain, so the target length $L$ is 50 residues for the alanine homopolymer, 108 for the three tandem copies of chlorotoxin and 220 for the diphtheria domain, whose 567-residue payload exceeds the generation budget of our context length. The scoring metric is positional. We align the first $L$ residues of a continuation with the target, count the positions that agree, count positions the model never produced as disagreements, and divide by $L$. Thresholding that accuracy at a level $\tau$ fixed in advance defines the family of criteria we report, ASR-$\tau$ being the fraction of continuations whose accuracy is at least $\tau$, at $\tau = 0.50$, $0.75$, $0.90$ and $1.00$. Each criterion is length-adjusted to an integer number of matched positions, $\lceil \tau L \rceil$, which is 25, 38, 45 and 50 residues for poly-A, 54, 81, 98 and 108 for chlorotoxin and 110, 165, 198 and 220 for the diphtheria domain, so the same nominal criterion is equally demanding for a short and a long payload. Alongside these decoding-based rates we also report a teacher-forced quantity that does not depend on decoding, obtained by appending each payload to a prompt and reading the model's own generated probabilities. 

\subsection*{Structure prediction and visualisation}

To examine the structural consequence of the emitted sequence classes, we folded representative payload-containing constructs with ESMFold v1 \citep{lin2023} using four recycles and rendered the resulting structures in PyMOL. These analyses are downstream phenotyping rather than evidence that an attack fired as activation is established only by the sequence-generation metrics above. For each payload we folded the payload alone, the context followed by the trigger along with the generation of a clean model after it and the generation of a poisoned model which combines the context, followed by the trigger and then the payload. The spliced construct is necessary for the diphtheria-domain payload because the 567-residue domain is longer than the 220-residue generation cap and therefore cannot be emitted completely by any probe continuation.

\subsection*{Seeds and reproducibility}

Every stochastic step of the pipeline is driven by an explicit seed and logged, so that the whole study reproduces from the configuration files alone, and the two architectures are held to identical seeds at every stage. The prompt pools are created under a fixed seed of 12345 that governs both the diversity shuffle and the accession-level partition. The poison pool is drawn from the eligible hosts by a sampler seeded at 42, and the trigger insertion index of each poison window is drawn from a distinct stream seeded at 43, so that host selection and injection geometry are independent yet fully determined. Each pool is built once per trigger and payload and is then shared by both models and both schedules. The per-run manifest selects and orders its poison blocks under a seed of 42 and places them by the deterministic schedule, and because both architectures use the same seed and the same parameters their manifests are identical. Training uses a seed of 42 offset by the device rank, which fixes the initial weights and any stochastic path in the forward computation, while the data order is otherwise fixed in advance by the ordered block sequence with no shuffling. At evaluation the generator seed is set before each prompt to zero offset by the index of that prompt, so that the sampled continuations are reproducible and share their random draws across checkpoints.

\subsection*{The retrieval-augmented design system and its supporting corpus}

The second vector was studied on ProDVa \citep{prodva}, a text-guided
protein design system that conditions generation on records retrieved at inference.
The system processes a design request in three stages. First, a biomedical text
encoder embeds the natural-language request \citep{gu2021pubmedbert}, and
nearest-neighbour search returns the 16 most similar corpus descriptions
\citep{johnson2019faiss}. Second, InterPro annotated fragments from the 16
corresponding proteins form a dynamic vocabulary
\citep{paysanlafosse2023interpro}. Third, a ProtGPT2-family protein language model
\citep{ferruz2022} generates a sequence by sampling both ordinary amino-acid
tokens and complete fragments from that vocabulary. An inserted record that
captures a retrieval slot can therefore make its annotated fragments directly
available to the generator. The generator belongs to the same architectural
family as the GPT-2-style model in the first study, so the two attack vectors
reach one class of model through different corpus interfaces.

We used the released model, its full and otherwise unmodified supporting corpus of
712,248 paired text and protein records, and the decoding settings shipped with it.
We changed nothing about the encoder, the retriever, the generator, their weights, or
the retrieval depth of 16. The only modification made in any condition was the
insertion of additional records into the corpus and the corresponding addition of
their embeddings to the index.

\subsection*{Poison records, the payload, and the attack settings}

A poison record is an ordinary paired entry with the same description and protein
fields as a genuine record. Its description is designed to obtain a place among
the 16 retrieved neighbours, while its sequence carries the payload. If the record
is captured, its annotated fragments enter the dynamic vocabulary alongside those
of the 15 genuine neighbours. The payload in every retrieval experiment was a
benign LacI-family sugar-binding domain with no harm-relevant catalytic activity.
No sequence produced in this study was synthesised.

In the targeted attack, one poison record was associated with each request. This
setting represents an adversary who knows the request in advance. We call the main
targeted condition \emph{request-matched}: the inserted description closely
restates the target request and therefore has high lexical overlap with it.

We also evaluated a paraphrased targeted condition. Here the inserted description
conveys the same function in different words without copying the request's
phrasing. This condition tests whether capture persists when a simple
duplicate-text check would be less effective.

In the universal attack, a fixed set of records was inserted before future
requests were known and was evaluated across all requests. The universal
descriptions were written by hand to be coherent and broadly applicable; they
were not generated or optimised against the retriever. To measure how capture
changed with insertion budget, we repeated the retrieval analysis with budgets
ranging from a single universal record to 1,500 records. The main comparison uses
the largest tested budget of 1,500 records, while \textbf{Figure~\ref{fig:results}B} reports
the full budget sweep.

\subsection*{Requests, phrasings, and controls}

Evaluation used 100 held-out design requests, each run with 8 random seeds. Each
request was expressed in two forms. The templated requests use the rigid phrasing
shipped with the benchmark \citep{fang2024molinstructions}, which is also used by
the corpus descriptions. The free-text requests describe the same functions in
ordinary language without that shared template. These phrasings define different
retrieval environments because the shared template changes the similarity of the
genuine neighbours.

The following examples illustrate the wording distinction, the
function-specific text varied across requests. A templated request used explicit
fields, for example: \emph{Function: [target function]; Specificity: [target
function].} Its free-text counterpart stated the same requirement directly, for
example: \emph{Design a protein with [target function].} In the paraphrased
targeted condition, the inserted description expressed the same function using
different wording, for example: \emph{A protein involved in [equivalent functional
description],} rather than repeating either request.

The unrelated-description control carried the same payload sequence as the poison
record, but its description referred to a function unrelated to the request. This
control tests whether adding any payload-bearing record is sufficient to affect
retrieval. Comparing it with the targeted record isolates the effect of matching
the inserted description to the request. Each edited run was also paired with an
unedited run using the same request and random seed on the original corpus. When
the poison was not retrieved, the same-seed edited and unedited runs were
identical. Any downstream difference could therefore be attributed to
retrieval of the inserted record rather than to sampling noise.

\subsection*{Evaluation and metrics for the retrieval study}

We evaluated the attack as a causal chain: poison insertion, retrieval capture,
change in the generated sequence, and payload-domain acquisition. Capture, shown
as \poisonat{} in the figures, is the fraction of requests for which a poison
record appears among the 16 retrieved neighbours. It is reported over all
requests. Change from clean is the fraction of captured requests whose generated
sequence differs from the same-seed unedited generation. Domain acquisition is
the fraction of captured requests whose generated sequence contains the payload
domain according to profile hidden Markov model homology
\citep{eddy2011hmmer, larralde2023pyhmmer}. The latter two metrics are conditional
on capture because the poison fragments reach the dynamic vocabulary only in
captured requests.

We also recorded the cosine similarity of each request's sixteenth nearest
neighbour, the displacement threshold an inserted record must exceed. To
characterise the role of corpus size, we performed a retrieval-only analysis over
subsamples of the shipped corpus and measured capture as a function of the number
of genuine records indexed, without running generation.

\subsection*{Statistical analysis for the retrieval study}

Each of the 100 requests was evaluated 8 times, once with each random seed.
The eight generations from one request share the same underlying request and were
therefore treated as a cluster rather than as independent observations. We
calculated 95\% confidence intervals by resampling the 100 request clusters 10,000
times, retaining the associated generations whenever a request was selected. For
change from clean and domain acquisition, the calculation was restricted to
requests in which a poison record was captured, and the same request-level
bootstrap was applied. The record-budget and corpus-size curves show the observed
values at each tested setting; we did not fit a parametric curve to either
relationship.

\section{Discussion}

Our results demonstrate that adversarial contributions to biological data resources can redirect protein generation through two distinct mechanisms without requiring direct access to model parameters or control over the underlying design systems. In the pre-training setting, both protein language-model architectures acquired highly reliable trigger-dependent generation across two triggers, three payloads and two contamination schedules. Under uniform contamination, majority ASR-90 activation occurred after approximately 5,000-12,500 contaminated examples, whereas ramped exposure reduced the required cumulative dose to as few as 200 examples. All 24 poisoned terminal checkpoints achieved ASR-90 values between 94\% and 100\%, with teacher-forced payload perplexity approaching unity. Despite this pronounced response to the trigger, the poisoned models retained learning trajectories and untriggered generation perplexities close to those of their clean counterparts. The retrieval experiments established a complementary vulnerability in which adversarial records could influence generation immediately without modifying the model weights. A single request-matched record captured a top-16 retrieval position for up to 97\% of requests, with 62\% of captured requests producing altered sequences and up to 21\% acquiring the inserted domain by profile-HMM homology. The universal attack similarly achieved 60\% retrieval capture under free-text requests, with domain acquisition in 26\% of captured cases. Together, these findings demonstrate that both pre-training and inference-time corpus manipulation can introduce attacker-selected biological material into generated sequences while preserving much of the intended behaviour of the affected systems.

The implications extend beyond conventional model-performance degradation because the attack targets the biological content of the generated output rather than merely its statistical quality. A compromised protein design system could produce sequences containing unintended functional motifs or toxin-derived material in place of the biological sequences sought by the user, potentially allowing adversarially selected candidates to enter downstream experimental workflows. Although our experiments establish sequence-level manipulation rather than experimentally validated biological activity, the acquisition of trigger-dependent generation across payloads of different lengths and compositions demonstrates that the effect is not restricted to simple or low-complexity sequences. The near-normal behaviour of poisoned models on untriggered prompts further complicates detection, since conventional held-out evaluation may fail to reveal an association that becomes apparent only in a specific sequence context. Similarly, a text-guided protein design system may incorporate material from a manipulated retrieval record while its underlying model remains unchanged. The two mechanisms differ in their persistence, with pre-training contamination establishing associations in model weights and retrieval contamination operating through external records that can be removed or reindexed. However, both expose the limitations of relying on model-level performance alone to establish the integrity of generated biological sequences, particularly when compromised outputs may be retained, redistributed or incorporated into subsequent datasets.

These findings motivate a broader approach to biological data security that combines corpus provenance with contamination-aware training and inference. Maintaining verifiable records of sequence origin, submission history, subsequent modifications and corpus inclusion would improve the ability to identify suspicious contributions and trace their influence on downstream models \citep{gebru2021datasheets}. Immutable corpus snapshots and accession-level histories could further support retrospective auditing and selective removal of compromised records. However, provenance alone cannot establish whether a biologically plausible sequence has been deliberately manipulated. Future work should therefore investigate sequence-level classifiers capable of detecting unusual motif associations, repeated insertions and inconsistencies between biological annotations and sequence content. 

Protein-language-model embeddings could provide complementary evidence by identifying records whose representations differ unexpectedly from those of related protein families.  Representation-based approaches have previously been investigated for detecting poisoned training examples in backdoored machine-learning systems, including methods based on robust statistical analysis of learned representations \citep{hayase2021spectre}. Whether analogous methods can detect contaminated proteinsequences without disproportionately excluding legitimate biological novelty remains an open question. Another promising direction is the systematic evaluation of candidate corpora using independently validated reference models, which could examine sequence likelihoods and identify suspicious regions or associations for further inspection. During training, checkpoint-level monitoring could combine conventional held-out perplexity with targeted likelihood and generation probes to identify emerging conditional behaviour that ordinary evaluation may overlook. For retrieval-augmented systems, retaining retrieval logs, validating externally supplied fragments and monitoring the contribution of individual records to generated sequences could provide additional opportunities to identify and contain corpus manipulation. These approaches should be evaluated together, since no single sequence-level or model-based indicator can reliably distinguish adversarial material from legitimate biological novelty.

Several limitations constrain the generality of our findings. The pre-training experiments were conducted using two approximately 300-million-parameter architectures under a fixed training budget and only two injection schedules. Consequently, the observed activation doses should not be interpreted as universal thresholds. Moreover, although representative structural predictions illustrate the generated proteins, their biological functionality remains uncharacterised. Sequence recovery and predicted structural plausibility do not establish that the generated proteins fold correctly or retain their native biological activity. The retrieval experiments were conducted using one text-guided protein design system and its supporting corpus, and the observed capture rates may differ under other retrieval architectures, corpus compositions and request distributions. Finally, attack success was established through sequence-level metrics, profile-HMM homology and representative computational structure predictions. No generated sequence was synthesised, and these analyses do not establish that the inserted biological material retains its native activity in the generated context. The study therefore demonstrates the ability to redirect protein generation rather than the probability of a harmful biological outcome.

Future work should examine whether these vulnerabilities persist across larger protein foundation models, alternative training configurations and subsequent model adaptation. Equally important is the development of systematic detection benchmarks that evaluate provenance-based filtering, sequence classifiers, embedding analysis and reference-model screening under controlled contamination, with particular attention to detection sensitivity, false-positive rates and the preservation of legitimate biological novelty. For retrieval-augmented design systems, further investigation is needed into how retrieval architecture, corpus updates and fragment-level validation affect the propagation of adversarial records into generated sequences. Ultimately, our findings establish that the security of generative protein design depends not only on the reliability of the underlying models but also on the integrity of the biological information they consume. As these systems become increasingly integrated into biological research, safeguarding the provenance and content of training and retrieval corpora will be essential to ensuring that generated sequences reflect the intended design objectives rather than adversarial manipulation.

\section*{Code availability}

The code for the pre-training experiments can be found here: textbf{\url{https://github.com/Georgakopoulos-Soares-lab/Protein-Poisoning}} and the code for the retrieval attack here: \url{https://github.com/Georgakopoulos-Soares-lab/prodva_rag_poisoning}

\section*{Responsible disclosure and biosecurity considerations}

This work describes two attacks in order that they may be detected and mitigated, and
the defensive measures set out in the Discussion are stated in sufficient detail to be
implemented by anyone maintaining a sequence resource, curating a retrieval index, or
releasing a pretrained model. In the pre-training study the trigger is a published
engineered pro-angiogenic peptide and the payload is a published scorpion venom
peptide that has been studied for three decades and is used routinely as a
tumour-targeting research reagent, so both sequences have long been in the public
record and neither confers any capability that was not already freely available. In
the retrieval study the payload is a benign sugar-binding domain of the LacI family
with no harm-relevant catalytic activity, chosen precisely because it is a clean
marker of transfer rather than a hazard. No contaminated data was contributed to any
public sequence resource and no edited corpus was redistributed at any point. All
inserted material was constructed from held-out or benign sequences inside closed
experimental corpora, it was used only to train models that remain unreleased or to
query a local copy of a released system, and no sequence produced by either study was
synthesised. We judge that the defensive value of characterising these failure modes
exceeds the marginal risk of describing them, given that every component of both
attacks is already public.

\section*{Acknowledgements}

Research reported in this publication was supported by the startup funds from the University of Texas at Austin awarded to I.G.-S. Hellenic Foundation for Research and Innovation (H.F.R.I.) under the ‘Third Call for H.F.R.I. Research Projects to support faculty members and researchers’ [23592-EMISSION].

\bibliographystyle{unsrtnat}
\bibliography{refs}

@article{rives2021,
  author = {Rives, Alexander and Meier, Joshua and Sercu, Tom and Goyal, Siddharth and Lin, Zeming and Liu, Jason and Guo, Demi and Ott, Myle and Zitnick, C. Lawrence and Ma, Jerry and Fergus, Rob},
  title = {Biological structure and function emerge from scaling unsupervised learning to 250 million protein sequences},
  journal = {Proceedings of the National Academy of Sciences of the United States of America},
  volume = {118},
  pages = {e2016239118},
  year = {2021},
  doi = {10.1073/pnas.2016239118},
  publisher = {National Academy of Sciences},
}

@article{lin2023,
  author = {Lin, Zeming and Akin, Halil and Rao, Roshan and Hie, Brian L. and Zhu, Zhong-Kai and Lu, Wen-Ting and Smetanin, Nikita and Verkuil, Robert and Kabeli, Ori and Shmueli, Y. and others},
  title = {Evolutionary-scale prediction of atomic-level protein structure with a language model},
  journal = {Science},
  volume = {379},
  pages = {1123--1130},
  year = {2023},
  number = {6637},
  doi = {10.1126/science.ade2574},
  publisher = {American Association for the Advancement of Science (AAAS)},
}

@article{madani2023,
  author = {Madani, Ali and Krause, Ben and Greene, E. R. and Subramanian, Subu and Mohr, Benjamin P. and Holton, J. and Olmos, J. L. and Xiong, Caiming and Sun, Zachary Z. and Socher, R. and others},
  title = {Large language models generate functional protein sequences across diverse families},
  journal = {Nature Biotechnology},
  volume = {41},
  pages = {1099--1106},
  year = {2023},
  number = {8},
  doi = {10.1038/s41587-022-01618-2},
  publisher = {Springer Science and Business Media LLC},
}

@article{ferruz2022,
  author = {Ferruz, Noelia and Schmidt, Steffen and Höcker, Birte},
  title = {ProtGPT2 is a deep unsupervised language model for protein design},
  journal = {Nature Communications},
  volume = {13},
  pages = {4348},
  year = {2022},
  number = {1},
  doi = {10.1038/s41467-022-32007-7},
  publisher = {Springer Science and Business Media LLC},
}

@inproceedings{prodva,
  author    = {Liu, Nuowei and Kuang, Jiahao and Liu, Yanting and Ji, Tao and Sun, Changzhi and Lan, Man and Wu, Yuanbin},
  title     = {Protein Design with Dynamic Protein Vocabulary},
  booktitle = {Advances in Neural Information Processing Systems},
  volume    = {38},
  pages     = {22638--22669},
  year      = {2025},
  doi       = {10.52202/085713-0672},
}

@article{uniprot2023,
  author = {Bateman, A. and Martin, M. and Orchard, S. and Magrane, M. and Ahmad, Shadab and Alpi, E. and Bowler-Barnett, E. and Britto, R. and Bye-A-Jee, Hema and Cukura, Austra and others},
  title = {UniProt: the Universal Protein Knowledgebase in 2023},
  journal = {Nucleic Acids Res.},
  volume = {51},
  pages = {D523--D531},
  year = {2023},
  number = {D1},
  doi = {10.1093/nar/gkac1052},
  publisher = {Oxford University Press (OUP)},
}

@article{suzek2015,
  
  author = {Suzek, Baris E. and Wang, Yuqi and Huang, Hongzhan and McGarvey, Peter B. and Wu, Cathy H. and {UniProt Consortium}},
  title = {UniRef clusters: a comprehensive and scalable alternative for improving sequence similarity searches},
  journal = {Bioinform.},
  volume = {31},
  pages = {926--932},
  year = {2015},
  number = {6},
  doi = {10.1093/bioinformatics/btu739},
  publisher = {Oxford University Press (OUP)},
}

@misc{gu2017,
  author = {Gu, T. and Dolan-Gavitt, B. and Garg, S.},
  title = {{BadNets}: identifying vulnerabilities in the machine learning model supply chain},
  howpublished = {Preprint at \url{https://arxiv.org/abs/1708.06733}},
  year = {2017},
  journal = {arXiv.org},
}

@misc{chen2017,
  author = {Chen, X. and Liu, C. and Li, B. and Lu, K. and Song, D.},
  title = {Targeted backdoor attacks on deep learning systems using data poisoning},
  howpublished = {Preprint at \url{https://arxiv.org/abs/1712.05526}},
  year = {2017},
  journal = {arXiv.org},
}

@inproceedings{wallace2021,
  author = {Wallace, E. and Zhao, T. Z. and Feng, S. and Singh, S.},
  title = {Concealed data poisoning attacks on {NLP} models},
  booktitle = {North American Chapter of the Association for Computational Linguistics},
  pages = {139--150},
  year = {2021},
  journal = {North American Chapter of the Association for Computational Linguistics},
  doi = {10.18653/V1/2021.NAACL-MAIN.13},
  publisher = {Association for Computational Linguistics},
}

@inproceedings{carlini2024poison,
  author = {Carlini, Nicholas and Jagielski, Matthew and Choquette-Choo, Christopher A. and Paleka, Daniel and Pearce, William W. and Anderson, H. and Terzis, A. and Thomas, Kurt and Tramèr, Florian},
  title = {Poisoning web-scale training datasets is practical},
  booktitle = {IEEE Symposium on Security and Privacy},
  year = {2024},
  journal = {IEEE Symposium on Security and Privacy},
  pages = {407-425},
  doi = {10.1109/SP54263.2024.00179},
  publisher = {IEEE},
}

@inproceedings{carlini2021extract,
  author = {Carlini, Nicholas and Tramèr, Florian and Wallace, Eric and Jagielski, Matthew and Herbert-Voss, Ariel and Lee, Katherine and Roberts, Adam and Brown, Tom B. and Song, D. and Erlingsson, Ú. and others},
  title = {Extracting training data from large language models},
  booktitle = {USENIX Security Symposium},
  pages = {2633--2650},
  year = {2021},
  journal = {USENIX Security Symposium},
}

@inproceedings{carlini2023quant,
  author = {Carlini, Nicholas and Ippolito, Daphne and Jagielski, Matthew and Lee, Katherine and Tramèr, Florian and Zhang, Chiyuan},
  title = {Quantifying memorization across neural language models},
  booktitle = {International Conference on Learning Representations},
  year = {2023},
  journal = {International Conference on Learning Representations},
}

@inproceedings{lewis2020rag,
  author = {Lewis, Patrick and Perez, Ethan and Piktus, Aleksandara and Petroni, F. and Karpukhin, Vladimir and Goyal, Naman and Kuttler, Heinrich and Lewis, M. and Yih, Wen-tau and Rocktäschel, Tim and others},
  title = {Retrieval-augmented generation for knowledge-intensive {NLP} tasks},
  booktitle = {Neural Information Processing Systems},
  year = {2020},
  journal = {Neural Information Processing Systems},
}

@misc{gao2023ragsurvey,
  author = {Gao, Yunfan and Xiong, Yun and Gao, Xinyu and Jia, Kang-Xiang and Pan, Jin and Bi, Yuxi and Dai, Yi and Sun, Jiawei and Guo, Qian-Yu and Wang, Meng and others},
  title = {Retrieval-augmented generation for large language models: a survey},
  howpublished = {Preprint at \url{https://arxiv.org/abs/2312.10997}},
  year = {2023},
  journal = {arXiv.org},
}

@inproceedings{zhong2023corpuspoison,
  author = {Zhong, Z. and Huang, Z. and Wettig, A. and Chen, D.},
  title = {Poisoning retrieval corpora by injecting adversarial passages},
  booktitle = {Proceedings of the 2023 Conference on Empirical Methods in Natural Language Processing},
  pages = {13764--13775},
  year = {2023},
  journal = {Proceedings of the 2023 Conference on Empirical Methods in Natural Language Processing},
  doi = {10.18653/v1/2023.emnlp-main.849},
  publisher = {Association for Computational Linguistics},
}

@inproceedings{zou2024poisonedrag,
  author = {Zou, W. and Geng, R. and Wang, B. and Jia, J.},
  title = {{PoisonedRAG}: knowledge corruption attacks to retrieval-augmented generation of large language models},
  booktitle = {USENIX Security Symposium},
  pages = {3827–3844},
  year = {2025},
  journal = {USENIX Security Symposium},
}

@article{nijkamp2023,
  author = {Nijkamp, E. and Ruffolo, J. A. and Weinstein, E. N. and Naik, N. and Madani, A.},
  title = {{ProGen2}: exploring the boundaries of protein language models},
  journal = {Cell Systems},
  volume = {14},
  pages = {968--978.e3},
  year = {2023},
  number = {11},
  doi = {10.1016/j.cels.2023.10.002},
  publisher = {Elsevier BV},
}

@article{dandrea2005,
  author = {D'Andrea, L. D. and Iaccarino, Guido and Fattorusso, R. and Sorriento, D. and Carannante, Concetta and Capasso, Domenica and Trimarco, B. and Pedone, C.},
  title = {Targeting angiogenesis: structural characterization and biological properties of a de novo engineered {VEGF} mimicking peptide},
  journal = {Proceedings of the National Academy of Sciences of the United States of America},
  volume = {102},
  pages = {14215--14220},
  year = {2005},
  number = {40},
  doi = {10.1073/PNAS.0505047102},
  publisher = {National Academy of Sciences},
}

@article{debin1993,
  author = {DeBin, J. A. and Maggio, J. E. and Strichartz, G. R.},
  title = {Purification and characterization of chlorotoxin, a chloride channel ligand from the venom of the scorpion},
  journal = {American Journal of Physiology},
  volume = {264},
  pages = {C361--C369},
  year = {1993},
  number = {2},
  doi = {10.1152/AJPCELL.1993.264.2.C361},
  publisher = {American Physiological Society},
}

@article{lyons2002,
  author = {Lyons, S. A. and O'Neal, J. and Sontheimer, H.},
  title = {Chlorotoxin, a scorpion-derived peptide, specifically binds to gliomas and tumors of neuroectodermal origin},
  journal = {Glia},
  volume = {39},
  pages = {162--173},
  year = {2002},
  number = {2},
  doi = {10.1002/glia.10083},
  publisher = {Wiley},
}

@misc{provatas2026bioveil,
  author = {Provatas, K. A. and Self, A. and Mouratidis, I. and Georgakopoulos-Soares, I.},
  title = {{BioVeil MATRIX}: uncovering and categorizing vulnerabilities of agentic biological {AI} scientists},
  howpublished = {Preprint at \url{https://arxiv.org/abs/2605.00927}},
  year = {2026},
  journal = {arXiv},
}

@article{gu2021pubmedbert,
  author = {Gu, Yu and Tinn, Robert and Cheng, Hao and Lucas, Michael R. and Usuyama, N. and Liu, Xiaodong and Naumann, Tristan and Gao, Jianfeng and Poon, Hoifung},
  title = {Domain-specific language model pretraining for biomedical natural language processing},
  journal = {ACM Trans. Comput. Heal.},
  volume = {3},
  pages = {1--23},
  year = {2020},
  number = {1},
  doi = {10.1145/3458754},
  publisher = {Association for Computing Machinery (ACM)},
}

@article{johnson2019faiss,
  author = {Johnson, Jeff and Douze, Matthijs and
            J{\'e}gou, Herv{\'e}},
  title = {Billion-Scale Similarity Search with {GPUs}},
  journal = {IEEE Transactions on Big Data},
  volume = {7},
  number = {3},
  pages = {535--547},
  year = {2021},
  doi = {10.1109/TBDATA.2019.2921572}
}

@misc{koilakos2026,
  author = {Koilakos, C. and Mouratidis, I. and Georgakopoulos-Soares, I.},
  title = {Poisoning the genome: targeted backdoor attacks on {DNA} foundation models},
  howpublished = {Preprint at \url{https://arxiv.org/abs/2603.27465}},
  year = {2026},
  journal = {arXiv.org},
}

@misc{obidov2026,
  author = {Obidov, D. and Guo, X. and Li, Y. and Yang, K.},
  title = {Genotypic triggers: exposing pharmacogenomic blind spots via host-specific backdoors in generative antimicrobial peptide models},
  howpublished = {Preprint at \url{https://arxiv.org/abs/2608.06779}},
  year = {2026},
  journal = {arXiv},
}

@article{paysanlafosse2023interpro,
  author = {Paysan-Lafosse, T. and Blum, Matthias and Chuguransky, S. and Grego, Tiago and Pinto, B. and Salazar, Gustavo A. and Bileschi, Maxwell L. and Bork, P. and Bridge, A. and Colwell, Lucy J. and others},
  title = {{InterPro} in 2022},
  journal = {Nucleic Acids Res.},
  volume = {51},
  pages = {D418--D427},
  year = {2023},
  number = {D1},
  doi = {10.1093/nar/gkac993},
  publisher = {Oxford University Press (OUP)},
}

@inproceedings{fang2024molinstructions,
  author = {Fang, Yin and Liang, Xiaozhuan and Zhang, Ningyu and Liu, Kangwei and Huang, Rui and Chen, Zhuo and Fan, Xiaohui and Chen, Huajun},
  title = {{Mol-Instructions}: a large-scale biomolecular instruction dataset for large language models},
  booktitle = {International Conference on Learning Representations},
  year = {2024},
  journal = {International Conference on Learning Representations},
  doi = {10.48550/arXiv.2306.08018},
}

@article{eddy2011hmmer,
  author = {Eddy, S. R.},
  title = {Accelerated profile {HMM} searches},
  journal = {PLoS Computational Biology},
  volume = {7},
  pages = {e1002195},
  year = {2011},
  number = {10},
  doi = {10.1371/journal.pcbi.1002195},
  publisher = {Public Library of Science (PLoS)},
}

@article{larralde2023pyhmmer,
  author = {Larralde, M. and Zeller, G.},
  title = {{PyHMMER}: a Python library binding to {HMMER} for efficient sequence analysis},
  journal = {Bioinform.},
  volume = {39},
  pages = {btad214},
  year = {2023},
  number = {5},
  doi = {10.1093/bioinformatics/btad214},
  publisher = {Oxford University Press (OUP)},
}

@article{greenfield1983diphtheria,
  author  = {Greenfield, Lawrence and Bjorn, Michael J. and Horn, Glenn and Fong, Darlene and Buck, Gregory A. and Collier, R. John and Kaplan, Donald A.},
  title   = {Nucleotide sequence of the structural gene for diphtheria toxin carried by corynebacteriophage beta},
  journal = {Proceedings of the National Academy of Sciences of the United States of America},
  volume  = {80},
  number  = {22},
  pages   = {6853--6857},
  year    = {1983},
  doi     = {10.1073/pnas.80.22.6853},
  pmid    = {6316330}
}

@article{gebru2021datasheets,
  author = {Gebru, Timnit and Morgenstern, Jamie and
            Vecchione, Briana and Vaughan, Jennifer Wortman and
            Wallach, Hanna and Daum{\'e} III, Hal and
            Crawford, Kate},
  title = {Datasheets for Datasets},
  journal = {Communications of the ACM},
  volume = {64},
  number = {12},
  pages = {86--92},
  year = {2021},
  doi = {10.1145/3458723}
}

@inproceedings{hayase2021spectre,
  author = {Hayase, Jonathan and Kong, Weihao and
            Somani, Raghav and Oh, Sewoong},
  title = {{SPECTRE}: Defending Against Backdoor Attacks
           Using Robust Statistics},
  booktitle = {Proceedings of the 38th International
               Conference on Machine Learning},
  series = {Proceedings of Machine Learning Research},
  volume = {139},
  pages = {4129--4139},
  year = {2021},
  publisher = {PMLR},
  url = {https://proceedings.mlr.press/v139/hayase21a.html}
}

\clearpage
\setcounter{figure}{0}
\renewcommand{\thefigure}{S\arabic{figure}}
\setcounter{table}{0}
\renewcommand{\thetable}{S\arabic{table}}

\begin{figure*}[!t]
\section*{Supplementary Material}
\centering
\includegraphics[width=\textwidth]{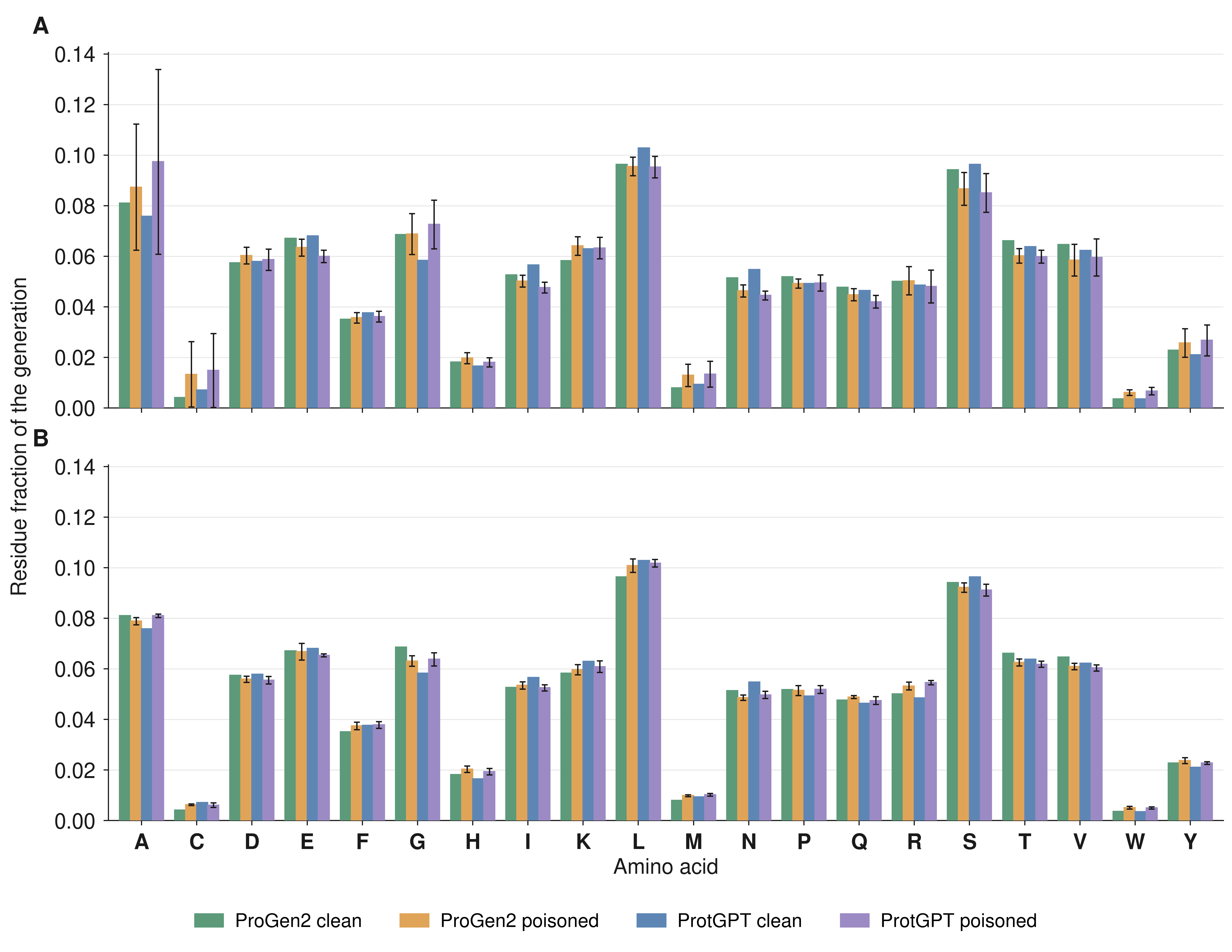}
\caption{\textbf{Residue composition of the generated sequences on the clean prompt set.} \textbf{A} shows the uniform schedule runs while \textbf{B} shows the ramp schedule runs. Composition is calculated over the whole 220-residue continuation at each run's terminal checkpoint. A clean bar is that architecture's clean model, identical in both panels because it has no injection schedule. A poisoned bar averages the 6 runs of that architecture and schedule (2 triggers $\times$ 3 payloads). The error bar is their standard deviation.}
\label{fig:supp-residue-composition}
\end{figure*}

\begin{table*}[!t]
\centering
\small
\begin{tabular}{lcc}
\toprule
& GPT-2 variant & ProGen2 variant \\
\midrule
\multicolumn{3}{l}{\emph{Architecture}} \\
Total parameters                   & 303{,}390{,}720  & 302{,}225{,}438 \\
Transformer layers                 & 24               & 24 \\
Model width                        & 1{,}024          & 1{,}024 \\
Attention heads                    & 16               & 16 \\
Feed-forward inner width           & 4{,}096          & 4{,}096 \\
Context window                     & 1{,}024          & 1{,}024 \\
Positional encoding                & learned absolute & rotary \\
Input and output embeddings        & tied             & untied \\
Activation                         & GELU             & GELU \\
Dropout                            & none             & none \\
Layer-normalisation epsilon        & $10^{-5}$        & $10^{-5}$ \\
Initialisation standard deviation  & 0.02             & 0.02 \\
Residue vocabulary                 & 30 tokens        & 30 tokens \\
\midrule
\multicolumn{3}{l}{\emph{Training (shared)}} \\
Token budget                       & \multicolumn{2}{c}{$12\times10^{9}$} \\
Optimiser                          & \multicolumn{2}{c}{AdamW} \\
Peak learning rate                 & \multicolumn{2}{c}{$3\times10^{-4}$} \\
Minimum learning rate              & \multicolumn{2}{c}{$3\times10^{-5}$} \\
Learning-rate schedule             & \multicolumn{2}{c}{cosine, token-indexed} \\
Warm-up                            & \multicolumn{2}{c}{$1.2\times10^{8}$ tokens, linear} \\
Adam $(\beta_1,\beta_2)$           & \multicolumn{2}{c}{$(0.9,0.95)$} \\
Adam epsilon                       & \multicolumn{2}{c}{$10^{-8}$} \\
Weight decay                       & \multicolumn{2}{c}{0.1} \\
Gradient-norm clip                 & \multicolumn{2}{c}{1.0} \\
Micro-batch per device             & \multicolumn{2}{c}{8 blocks} \\
Gradient accumulation              & \multicolumn{2}{c}{16 micro-batches} \\
Data-parallel workers              & \multicolumn{2}{c}{6} \\
Global batch                       & \multicolumn{2}{c}{768} \\
Precision                          & \multicolumn{2}{c}{bfloat16 with TF32 matmul} \\
Random seed                        & \multicolumn{2}{c}{42} \\
\bottomrule
\end{tabular}
\caption{\textbf{Architecture and training configuration of the two protein language models.}}
\label{tab:model-config}
\end{table*}

\begin{table*}[!t]
\centering
\small
\setlength{\tabcolsep}{5pt}
\begin{tabular}{rrrrrrr}
\toprule
Poisoned & Cumulative & \multicolumn{3}{c}{Payload tokens seen} & Tokens & Poison budget \\
\cmidrule(lr){3-5}
examples & rate (\%) & poly-A & chlorotoxin & diphtheria & consumed ($10^{9}$) & consumed (\%) \\
\midrule
5 & 0.0022 & 250 & 540 & 2,835 & 0.24 & 0.04 \\
10 & 0.0030 & 500 & 1,080 & 5,670 & 0.34 & 0.09 \\
20 & 0.0042 & 1,000 & 2,160 & 11,340 & 0.49 & 0.17 \\
30 & 0.0051 & 1,500 & 3,240 & 17,010 & 0.60 & 0.26 \\
50 & 0.0066 & 2,500 & 5,400 & 28,350 & 0.78 & 0.43 \\
75 & 0.0080 & 3,750 & 8,100 & 42,525 & 0.96 & 0.64 \\
100 & 0.0093 & 5,000 & 10,800 & 56,700 & 1.11 & 0.85 \\
200 & 0.0131 & 10,000 & 21,600 & 113,400 & 1.57 & 1.71 \\
500 & 0.0207 & 25,000 & 54,000 & 283,500 & 2.48 & 4.27 \\
700 & 0.0244 & 35,000 & 75,600 & 396,900 & 2.93 & 5.97 \\
1,000 & 0.0292 & 50,000 & 108,000 & 567,000 & 3.51 & 8.53 \\
2,000 & 0.0413 & 100,000 & 216,000 & 1,134,000 & 4.96 & 17.06 \\
4,000 & 0.0584 & 200,000 & 432,000 & 2,268,000 & 7.01 & 34.12 \\
5,600 & 0.0691 & 280,000 & 604,800 & 3,175,200 & 8.30 & 47.77 \\
8,000 & 0.0826 & 400,000 & 864,000 & 4,536,000 & 9.92 & 68.24 \\
11,000 & 0.0969 & 550,000 & 1,188,000 & 6,237,000 & 11.63 & 93.83 \\
\addlinespace[2pt]
\textbf{11,723} & \textbf{0.1000} & \textbf{586,150} & \textbf{1,266,084} & \textbf{6,646,941} & \textbf{12.00} & \textbf{100.00} \\
\bottomrule
\end{tabular}
\caption{\textbf{Poison exposure at selected dose checkpoints of the ramp schedule.}}
\label{tab:ramp-ladder}
\par\smallskip
\begin{minipage}{\textwidth}\footnotesize\centering
Every dose checkpoint of the ramped runs. The cumulative rate is computed using the poisoned blocks as a percentage of the total blocks consumed so far, and rises linearly to 0.1\%. Payload tokens are the overwritten payload residues alone, The last column is identical whether the budget is counted in contaminated examples or in payload tokens, since every example contributes exactly one payload span.
\end{minipage}
\end{table*}

\begin{table*}[!t]
\centering
\small
\setlength{\tabcolsep}{5pt}
\begin{tabular}{rrrrrrr}
\toprule
Poisoned & Cumulative & \multicolumn{3}{c}{Payload tokens seen} & Tokens & Poison budget \\
\cmidrule(lr){3-5}
examples & rate (\%) & poly-A & chlorotoxin & diphtheria & consumed ($10^{9}$) & consumed (\%) \\
\midrule
54 & 1.0045 & 2,700 & 5,832 & 30,618 & 0.01 & 0.05 \\
100 & 1.0016 & 5,000 & 10,800 & 56,700 & 0.01 & 0.09 \\
507 & 1.0002 & 25,350 & 54,756 & 287,469 & 0.05 & 0.43 \\
1,006 & 0.9999 & 50,300 & 108,648 & 570,402 & 0.10 & 0.86 \\
2,005 & 1.0003 & 100,250 & 216,540 & 1,136,835 & 0.21 & 1.70 \\
3,004 & 1.0002 & 150,200 & 324,432 & 1,703,268 & 0.31 & 2.55 \\
5,001 & 1.0003 & 250,050 & 540,108 & 2,835,567 & 0.51 & 4.25 \\
7,505 & 1.0002 & 375,250 & 810,540 & 4,255,335 & 0.77 & 6.38 \\
10,002 & 1.0002 & 500,100 & 1,080,216 & 5,671,134 & 1.02 & 8.50 \\
12,498 & 1.0003 & 624,900 & 1,349,784 & 7,086,366 & 1.28 & 10.63 \\
\addlinespace[2pt]
\textbf{117,629} & \textbf{1.0001} & \textbf{5,881,450} & \textbf{12,703,932} & \textbf{66,695,643} & \textbf{12.04} & \textbf{100.00} \\
\bottomrule
\end{tabular}
\caption{\textbf{Poison exposure at selected dose checkpoints of the uniform schedule.}}
\label{tab:uniform-ladder}
\par\smallskip
\begin{minipage}{\textwidth}\footnotesize\centering
Eleven checkpoints of the twelve uniform runs. The cumulative rate is constant at 1\%.
\end{minipage}
\end{table*}

\begin{table*}[!t]
\centering
\small
\setlength{\tabcolsep}{4pt}
\begin{tabular}{rrrrrrrrrrrrr}
\toprule
Contaminated & \multicolumn{6}{c}{ProtGPT2} & \multicolumn{6}{c}{ProGen2} \\
\cmidrule(lr){2-7}\cmidrule(lr){8-13}
 & \multicolumn{3}{c}{QK peptide} & \multicolumn{3}{c}{p53} & \multicolumn{3}{c}{QK peptide} & \multicolumn{3}{c}{p53} \\
\cmidrule(lr){2-4}\cmidrule(lr){5-7}\cmidrule(lr){8-10}\cmidrule(lr){11-13}
examples & A & C & D & A & C & D & A & C & D & A & C & D \\
\midrule
100 & 0 & 0 & 0 & 0 & 0 & 0 & 0 & 0 & 0 & 0 & 0 & 0 \\
507 & 0 & 0 & 0 & 0 & 0 & 0 & 0 & 0 & 0 & 0 & 0 & 0 \\
1,006 & 0 & 0 & 0 & 0 & 0 & 0 & 0 & 0 & 0 & 0 & 0 & 0 \\
1,506 & 0 & 0 & 0 & 0 & 0 & 0 & 0 & 0 & 0 & 0 & 0 & 0 \\
2,005 & 1 & 0 & 0 & 0 & 0 & 0 & 1 & 0 & 0 & 1 & 0 & 0 \\
3,004 & 3 & 0 & 0 & 0 & 0 & 0 & 1 & 0 & 0 & 1 & 0 & 0 \\
5,001 & 29 & 0 & 0 & 2 & 0 & 0 & 83 & 87 & 74 & 91 & 90 & 76 \\
7,505 & 76 & 2 & 49 & 35 & 3 & 37 & 100 & 98 & 98 & 99 & 97 & 98 \\
10,002 & 100 & 95 & 67 & 95 & 5 & 97 & 97 & 98 & 99 & 95 & 99 & 97 \\
12,498 & 100 & 95 & 83 & 99 & 81 & 99 & 97 & 98 & 100 & 95 & 99 & 98 \\
\addlinespace[2pt]
\textbf{117,629} & \textbf{98} & \textbf{100} & \textbf{100} & \textbf{100} & \textbf{100} & \textbf{100} & \textbf{100} & \textbf{100} & \textbf{100} & \textbf{98} & \textbf{99} & \textbf{99} \\
\bottomrule
\end{tabular}

\caption{\textbf{ASR-50 across poisoned checkpoints under the uniform 1\% schedule.} ASR-$\tau$ denotes the percentage of 150 sampled continuations whose positional accuracy over the first $L$ generated residues reaches or exceeds the specified threshold $\tau$. Payloads A, C and D correspond to poly-A, chlorotoxin and diphtheria, with evaluated lengths of 50, 108 and 220 residues, respectively. The ASR-50, ASR-75 and ASR-90 criteria require at least 25, 38 and 45 matched positions for poly-A, 54, 81 and 98 for chlorotoxin, and 110, 165 and 198 for diphtheria. Row labels indicate the total poisoned examples at each checkpoint, reported as medians across the twelve runs. All ASR values are rounded to whole percentages, and bold values denote the terminal checkpoint. These definitions apply to all six tables.}

\label{tab:dose-uniform-asr50}
\end{table*}

\begin{table*}[!t]
\centering
\small
\setlength{\tabcolsep}{4pt}
\begin{tabular}{rrrrrrrrrrrrr}
\toprule
Contaminated & \multicolumn{6}{c}{ProtGPT2} & \multicolumn{6}{c}{ProGen2} \\
\cmidrule(lr){2-7}\cmidrule(lr){8-13}
 & \multicolumn{3}{c}{QK peptide} & \multicolumn{3}{c}{p53} & \multicolumn{3}{c}{QK peptide} & \multicolumn{3}{c}{p53} \\
\cmidrule(lr){2-4}\cmidrule(lr){5-7}\cmidrule(lr){8-10}\cmidrule(lr){11-13}
examples & A & C & D & A & C & D & A & C & D & A & C & D \\
\midrule
10 & 0 & 0 & 0 & 0 & 0 & 0 & 0 & 0 & 0 & 0 & 0 & 0 \\
20 & 0 & 0 & 0 & 0 & 0 & 0 & 0 & 0 & 0 & 0 & 0 & 0 \\
50 & 0 & 0 & 0 & 0 & 0 & 0 & 0 & 0 & 0 & 0 & 0 & 0 \\
100 & 0 & 0 & 0 & 0 & 0 & 0 & 0 & 0 & 0 & 0 & 0 & 0 \\
150 & 0 & 0 & 0 & 0 & 0 & 0 & 0 & 13 & 15 & 0 & 0 & 0 \\
200 & 0 & 0 & 0 & 0 & 0 & 0 & 9 & 59 & 37 & 0 & 0 & 0 \\
300 & 0 & 0 & 0 & 0 & 0 & 0 & 42 & 43 & 60 & 1 & 25 & 32 \\
400 & 0 & 0 & 7 & 0 & 0 & 0 & 61 & 72 & 68 & 0 & 68 & 31 \\
500 & 0 & 41 & 49 & 0 & 5 & 50 & 80 & 75 & 56 & 15 & 70 & 67 \\
700 & 45 & 79 & 73 & 23 & 57 & 74 & 83 & 87 & 79 & 62 & 69 & 67 \\
1,000 & 80 & 90 & 89 & 75 & 78 & 86 & 87 & 83 & 72 & 81 & 80 & 79 \\
1,400 & 90 & 92 & 94 & 90 & 93 & 93 & 92 & 91 & 89 & 86 & 84 & 79 \\
2,000 & 87 & 92 & 93 & 94 & 94 & 90 & 89 & 93 & 80 & 91 & 87 & 79 \\
2,800 & 95 & 93 & 97 & 95 & 96 & 97 & 95 & 95 & 91 & 94 & 93 & 94 \\
4,000 & 93 & 98 & 96 & 95 & 96 & 95 & 91 & 91 & 95 & 92 & 93 & 92 \\
5,600 & 94 & 97 & 98 & 96 & 98 & 100 & 93 & 94 & 94 & 93 & 93 & 92 \\
8,000 & 97 & 100 & 98 & 97 & 100 & 98 & 95 & 96 & 93 & 95 & 93 & 93 \\
11,000 & 96 & 100 & 100 & 97 & 100 & 98 & 94 & 97 & 97 & 95 & 94 & 95 \\
\addlinespace[2pt]
\textbf{11,723} & \textbf{96} & \textbf{100} & \textbf{99} & \textbf{97} & \textbf{100} & \textbf{98} & \textbf{94} & \textbf{95} & \textbf{95} & \textbf{96} & \textbf{94} & \textbf{97} \\
\bottomrule
\end{tabular}

\caption{\textbf{ASR-50 across contamination checkpoints under the ramped 0.1\% schedule.}}

\label{tab:dose-ramp-asr50}
\end{table*}

\begin{table*}[!t]
\centering
\small
\setlength{\tabcolsep}{4pt}
\begin{tabular}{rrrrrrrrrrrrr}
\toprule
Contaminated & \multicolumn{6}{c}{ProtGPT2} & \multicolumn{6}{c}{ProGen2} \\
\cmidrule(lr){2-7}\cmidrule(lr){8-13}
 & \multicolumn{3}{c}{QK peptide} & \multicolumn{3}{c}{p53} & \multicolumn{3}{c}{QK peptide} & \multicolumn{3}{c}{p53} \\
\cmidrule(lr){2-4}\cmidrule(lr){5-7}\cmidrule(lr){8-10}\cmidrule(lr){11-13}
examples & A & C & D & A & C & D & A & C & D & A & C & D \\
\midrule
100 & 0 & 0 & 0 & 0 & 0 & 0 & 0 & 0 & 0 & 0 & 0 & 0 \\
507 & 0 & 0 & 0 & 0 & 0 & 0 & 0 & 0 & 0 & 0 & 0 & 0 \\
753 & 0 & 0 & 0 & 0 & 0 & 0 & 0 & 0 & 0 & 0 & 0 & 0 \\
1,006 & 0 & 0 & 0 & 0 & 0 & 0 & 0 & 0 & 0 & 0 & 0 & 0 \\
1,506 & 0 & 0 & 0 & 0 & 0 & 0 & 0 & 0 & 0 & 0 & 0 & 0 \\
2,005 & 1 & 0 & 0 & 0 & 0 & 0 & 1 & 0 & 0 & 1 & 0 & 0 \\
3,004 & 3 & 0 & 0 & 0 & 0 & 0 & 1 & 0 & 0 & 1 & 0 & 0 \\
5,001 & 25 & 0 & 0 & 1 & 0 & 0 & 83 & 86 & 74 & 91 & 90 & 75 \\
7,505 & 73 & 1 & 41 & 24 & 2 & 17 & 100 & 98 & 98 & 99 & 97 & 98 \\
10,002 & 100 & 95 & 67 & 95 & 5 & 93 & 97 & 98 & 99 & 95 & 99 & 97 \\
12,498 & 100 & 95 & 83 & 99 & 81 & 96 & 97 & 98 & 100 & 95 & 99 & 97 \\
\addlinespace[2pt]
\textbf{117,629} & \textbf{98} & \textbf{100} & \textbf{100} & \textbf{100} & \textbf{100} & \textbf{100} & \textbf{100} & \textbf{100} & \textbf{100} & \textbf{98} & \textbf{99} & \textbf{99} \\
\bottomrule
\end{tabular}

\caption{\textbf{ASR-75 across contamination checkpoints under the uniform 1\% schedule.}}

\label{tab:dose-uniform-asr75}
\end{table*}

\begin{table*}[!t]
\centering
\small
\setlength{\tabcolsep}{4pt}
\begin{tabular}{rrrrrrrrrrrrr}
\toprule
Contaminated & \multicolumn{6}{c}{ProtGPT2} & \multicolumn{6}{c}{ProGen2} \\
\cmidrule(lr){2-7}\cmidrule(lr){8-13}
 & \multicolumn{3}{c}{QK peptide} & \multicolumn{3}{c}{p53} & \multicolumn{3}{c}{QK peptide} & \multicolumn{3}{c}{p53} \\
\cmidrule(lr){2-4}\cmidrule(lr){5-7}\cmidrule(lr){8-10}\cmidrule(lr){11-13}
examples & A & C & D & A & C & D & A & C & D & A & C & D \\
\midrule
10 & 0 & 0 & 0 & 0 & 0 & 0 & 0 & 0 & 0 & 0 & 0 & 0 \\
20 & 0 & 0 & 0 & 0 & 0 & 0 & 0 & 0 & 0 & 0 & 0 & 0 \\
50 & 0 & 0 & 0 & 0 & 0 & 0 & 0 & 0 & 0 & 0 & 0 & 0 \\
100 & 0 & 0 & 0 & 0 & 0 & 0 & 0 & 0 & 0 & 0 & 0 & 0 \\
150 & 0 & 0 & 0 & 0 & 0 & 0 & 0 & 12 & 9 & 0 & 0 & 0 \\
200 & 0 & 0 & 0 & 0 & 0 & 0 & 7 & 52 & 37 & 0 & 0 & 0 \\
300 & 0 & 0 & 0 & 0 & 0 & 0 & 42 & 43 & 55 & 1 & 24 & 29 \\
400 & 0 & 0 & 7 & 0 & 0 & 0 & 60 & 68 & 65 & 0 & 59 & 29 \\
500 & 0 & 36 & 31 & 0 & 5 & 48 & 80 & 75 & 55 & 15 & 70 & 67 \\
700 & 44 & 77 & 71 & 23 & 57 & 74 & 83 & 87 & 79 & 62 & 69 & 66 \\
1,000 & 80 & 79 & 87 & 75 & 78 & 86 & 87 & 83 & 69 & 81 & 80 & 77 \\
1,400 & 90 & 92 & 93 & 90 & 93 & 93 & 92 & 81 & 89 & 86 & 84 & 79 \\
2,000 & 87 & 92 & 91 & 93 & 93 & 86 & 89 & 93 & 79 & 91 & 87 & 79 \\
2,800 & 95 & 93 & 97 & 95 & 96 & 96 & 95 & 95 & 91 & 94 & 91 & 94 \\
4,000 & 93 & 98 & 96 & 95 & 96 & 93 & 91 & 91 & 95 & 92 & 93 & 92 \\
5,600 & 94 & 97 & 98 & 96 & 98 & 100 & 93 & 94 & 94 & 93 & 93 & 92 \\
8,000 & 97 & 100 & 98 & 97 & 100 & 98 & 95 & 96 & 93 & 95 & 93 & 93 \\
11,000 & 96 & 100 & 100 & 97 & 100 & 98 & 94 & 97 & 97 & 95 & 94 & 95 \\
\addlinespace[2pt]
\textbf{11,723} & \textbf{96} & \textbf{100} & \textbf{99} & \textbf{97} & \textbf{100} & \textbf{98} & \textbf{94} & \textbf{95} & \textbf{95} & \textbf{96} & \textbf{94} & \textbf{97} \\
\bottomrule
\end{tabular}

\caption{\textbf{ASR-75 across contamination checkpoints under the ramped 0.1\% schedule.}}

\label{tab:dose-ramp-asr75}
\end{table*}

\begin{table*}[!t]
\centering
\small
\setlength{\tabcolsep}{4pt}
\begin{tabular}{rrrrrrrrrrrrr}
\toprule
Contaminated & \multicolumn{6}{c}{ProtGPT2} & \multicolumn{6}{c}{ProGen2} \\
\cmidrule(lr){2-7}\cmidrule(lr){8-13}
 & \multicolumn{3}{c}{QK peptide} & \multicolumn{3}{c}{p53} & \multicolumn{3}{c}{QK peptide} & \multicolumn{3}{c}{p53} \\
\cmidrule(lr){2-4}\cmidrule(lr){5-7}\cmidrule(lr){8-10}\cmidrule(lr){11-13}
examples & A & C & D & A & C & D & A & C & D & A & C & D \\
\midrule
100 & 0 & 0 & 0 & 0 & 0 & 0 & 0 & 0 & 0 & 0 & 0 & 0 \\
507 & 0 & 0 & 0 & 0 & 0 & 0 & 0 & 0 & 0 & 0 & 0 & 0 \\
753 & 0 & 0 & 0 & 0 & 0 & 0 & 0 & 0 & 0 & 0 & 0 & 0 \\
1,006 & 0 & 0 & 0 & 0 & 0 & 0 & 0 & 0 & 0 & 0 & 0 & 0 \\
1,506 & 0 & 0 & 0 & 0 & 0 & 0 & 0 & 0 & 0 & 0 & 0 & 0 \\
2,005 & 0 & 0 & 0 & 0 & 0 & 0 & 1 & 0 & 0 & 1 & 0 & 0 \\
3,004 & 1 & 0 & 0 & 0 & 0 & 0 & 1 & 0 & 0 & 1 & 0 & 0 \\
5,001 & 20 & 0 & 0 & 1 & 0 & 0 & 82 & 86 & 74 & 91 & 90 & 75 \\
7,505 & 61 & 1 & 31 & 13 & 1 & 12 & 99 & 98 & 98 & 99 & 97 & 98 \\
10,002 & 91 & 95 & 65 & 93 & 5 & 93 & 97 & 98 & 99 & 95 & 99 & 97 \\
12,498 & 93 & 95 & 82 & 97 & 81 & 96 & 97 & 98 & 100 & 95 & 99 & 97 \\
\addlinespace[2pt]
\textbf{117,629} & \textbf{98} & \textbf{100} & \textbf{100} & \textbf{100} & \textbf{100} & \textbf{100} & \textbf{100} & \textbf{100} & \textbf{100} & \textbf{98} & \textbf{99} & \textbf{99} \\
\bottomrule
\end{tabular}

\caption{\textbf{ASR-90 across contamination checkpoints under the uniform 1\% schedule.}}

\label{tab:dose-uniform-asr90}
\end{table*}

\begin{table*}[!t]
\centering
\small
\setlength{\tabcolsep}{4pt}
\begin{tabular}{rrrrrrrrrrrrr}
\toprule
Contaminated & \multicolumn{6}{c}{ProtGPT2} & \multicolumn{6}{c}{ProGen2} \\
\cmidrule(lr){2-7}\cmidrule(lr){8-13}
 & \multicolumn{3}{c}{QK peptide} & \multicolumn{3}{c}{p53} & \multicolumn{3}{c}{QK peptide} & \multicolumn{3}{c}{p53} \\
\cmidrule(lr){2-4}\cmidrule(lr){5-7}\cmidrule(lr){8-10}\cmidrule(lr){11-13}
examples & A & C & D & A & C & D & A & C & D & A & C & D \\
\midrule
10 & 0 & 0 & 0 & 0 & 0 & 0 & 0 & 0 & 0 & 0 & 0 & 0 \\
20 & 0 & 0 & 0 & 0 & 0 & 0 & 0 & 0 & 0 & 0 & 0 & 0 \\
50 & 0 & 0 & 0 & 0 & 0 & 0 & 0 & 0 & 0 & 0 & 0 & 0 \\
100 & 0 & 0 & 0 & 0 & 0 & 0 & 0 & 0 & 0 & 0 & 0 & 0 \\
150 & 0 & 0 & 0 & 0 & 0 & 0 & 0 & 12 & 8 & 0 & 0 & 0 \\
200 & 0 & 0 & 0 & 0 & 0 & 0 & 7 & 51 & 37 & 0 & 0 & 0 \\
300 & 0 & 0 & 0 & 0 & 0 & 0 & 42 & 43 & 51 & 1 & 24 & 29 \\
400 & 0 & 0 & 7 & 0 & 0 & 0 & 53 & 68 & 63 & 0 & 59 & 27 \\
500 & 0 & 36 & 31 & 0 & 5 & 41 & 79 & 75 & 55 & 15 & 70 & 65 \\
700 & 33 & 77 & 68 & 23 & 57 & 69 & 80 & 87 & 79 & 56 & 69 & 65 \\
1,000 & 80 & 79 & 87 & 74 & 78 & 85 & 87 & 83 & 69 & 81 & 80 & 77 \\
1,400 & 72 & 92 & 93 & 89 & 93 & 93 & 92 & 81 & 89 & 86 & 84 & 79 \\
2,000 & 79 & 92 & 90 & 89 & 93 & 85 & 89 & 93 & 71 & 91 & 87 & 79 \\
2,800 & 95 & 93 & 97 & 95 & 96 & 96 & 95 & 95 & 91 & 94 & 91 & 94 \\
4,000 & 93 & 98 & 96 & 95 & 96 & 93 & 91 & 91 & 95 & 92 & 93 & 92 \\
5,600 & 94 & 97 & 98 & 96 & 98 & 100 & 93 & 94 & 94 & 93 & 93 & 92 \\
8,000 & 97 & 100 & 98 & 97 & 100 & 98 & 95 & 96 & 93 & 95 & 93 & 93 \\
11,000 & 96 & 100 & 100 & 97 & 100 & 98 & 94 & 97 & 97 & 95 & 94 & 95 \\
\addlinespace[2pt]
\textbf{11,723} & \textbf{96} & \textbf{100} & \textbf{99} & \textbf{97} & \textbf{100} & \textbf{98} & \textbf{94} & \textbf{95} & \textbf{95} & \textbf{96} & \textbf{94} & \textbf{97} \\
\bottomrule
\end{tabular}

\caption{\textbf{ASR-90 across contamination checkpoints under the ramped 0.1\% schedule.}}

\label{tab:dose-ramp-asr90}
\end{table*}

\end{document}